\documentclass{article}

\usepackage{microtype}
\usepackage{graphicx}
\usepackage{subfigure}
\usepackage{booktabs} 
\usepackage[colorlinks=true, allcolors=blue]{hyperref}

\usepackage{CJKutf8}

\usepackage[accepted]{icml_r2fm}

\usepackage{extpfeil}
\usepackage{bm}
\usepackage{graphicx}
\usepackage{amsfonts}
\usepackage{amsmath}
\usepackage{amssymb}
\usepackage{mathtools}
\usepackage{amsthm}
\usepackage{algorithm}
\usepackage{algorithmic}
\usepackage{enumerate}
\usepackage{thmtools}
\usepackage[parfill]{parskip}
\theoremstyle{plain}
\usepackage[bb=ams]{mathalpha}
\newtheorem{theorem}{Theorem}[section]
\newtheorem{proposition}[theorem]{Proposition}
\newtheorem{lemma}[theorem]{Lemma}

\theoremstyle{definition}
\newtheorem{definition}[theorem]{Definition}

\theoremstyle{remark}

\DeclarePairedDelimiterX{\norm}[1]{\lVert}{\rVert}{#1}

\usepackage{amsmath}
\usepackage{graphicx}
\usepackage[colorlinks=true, allcolors=blue]{hyperref}
\usepackage{amsfonts}
\usepackage{caption}
\usepackage{subcaption}
\usepackage{extpfeil}
\usepackage{algorithm}
\usepackage{algorithmic}
\usepackage{tikz-cd}
\usepackage{pgfplots}
\usepackage{svg}
\usepackage{mathrsfs}

\usepackage{CJKutf8}
\usepackage{xcolor}

\definecolor{pc0}{rgb}{0, 0, 0}   
\definecolor{pc1}{rgb}{1, 0, 0}   
\definecolor{pc2}{rgb}{1, 0.6, 0}   
\definecolor{pc3}{rgb}{0.8, 0.8, 0}   
\definecolor{pc4}{rgb}{0.2, 1, 0}   
\definecolor{pc5}{rgb}{0, 0.8, 0.2}   
\definecolor{pc6}{rgb}{0, 0.6, 0.6}   
\definecolor{pc7}{rgb}{0, 0, 1}   
\definecolor{pc8}{rgb}{0.2, 0, 0.8}   

\icmltitlerunning{A Theoretical Framework for OOD Robustness in Transformers using Gevrey Classes}

\begin{document}

\twocolumn[
\icmltitle{A Theoretical Framework for OOD Robustness in Transformers\\ using Gevrey Classes}




\icmlsetsymbol{equal}{*}

\begin{icmlauthorlist}
\icmlauthor{Yu Wang}{equal,ntu}
\icmlauthor{Fu-Chieh Chang}{equal,ntu,media}
\icmlauthor{Pei-Yuan Wu}{ntu}
\end{icmlauthorlist}

\icmlaffiliation{ntu}{Graduate Institute of Communication Engineering, National Taiwan University, Taipei, Taiwan}
\icmlaffiliation{media}{MediaTek Research, Taipei, Taiwan}

\icmlcorrespondingauthor{Fu-Chieh Chang}{d09942015@ntu.edu.tw}

\icmlkeywords{chain-of-thought,large language models}

\vskip 0.3in
]



\printAffiliationsAndNotice{\icmlEqualContribution} 

\begin{abstract}
We study the robustness of Transformer language models under semantic out-of-distribution (OOD) shifts, where training and test data lie in disjoint latent spaces. Using Wasserstein-1 distance and Gevrey-class smoothness, we derive sub-exponential upper bounds on prediction error. Our theoretical framework explains how smoothness governs generalization under distributional drift. We validate these findings through controlled experiments on arithmetic and Chain-of-Thought tasks with latent permutations and scalings. Results show empirical degradation aligns with our bounds, highlighting the geometric and functional principles underlying OOD generalization in Transformers.
\end{abstract}

\section{Introduction}
\label{sec:introduction}

Large language models (LLMs) have shown remarkable success across a wide range of natural language processing (NLP) tasks~\cite{brown2020language,raffel2020exploring,achiam2023gpt,touvron2023llama,zhang2022opt}. Yet their behavior under distribution shift—especially when the training and testing data lie in semantically disjoint regions—remains poorly understood. Existing theoretical frameworks for generalization under distribution shift often assume overlapping support between the training and testing distributions, relying on divergences such as KL or domain discrepancy metrics~\cite{ben2010theory,germain2013pac,redko2017theoretical}. However, these divergences become ill-defined or uninformative when the support sets are disjoint, as they require absolute continuity. In realistic deployment scenarios, LLMs frequently encounter tasks that cannot be easily mapped onto their training distribution~\cite{arjovsky2019invariant,sagawa2019distributionally,volpi2018generalizing}, prompting the need for alternative measures. To this end, we adopt the Wasserstein-1 distance, which quantifies the geometric transport cost between distributions and remains well-defined even in disjoint settings.

In this work, we study the generalization behavior of Transformer-based sequence models under \textbf{latent semantic shift}, where the training and testing distributions are supported on non-overlapping subsets of the latent task space. We introduce a geometric and distributional framework for quantifying generalization degradation via the Wasserstein-1 distance, and derive new upper bounds on the mean squared error (MSE) of model predictions under such shift. Central to our analysis is the observation that autoregressive LLMs define function classes within a Gevrey class, whose controlled smoothness properties enable quantifiable robustness to structured input perturbations~\cite{opschoor2024exponential}. Our theoretical results are validated through a series of controlled experiments on simplified arithmetic and reasoning tasks, where the latent generative structure is explicitly known. In our experiment, 
we observe test-time degradation that closely follows the predicted sub-exponential upper bounds, reinforcing similar patterns observed in adversarial robustness~\cite{hein2017formal,zhang2019theoretically}.

In addition to these main results, we also extend an existing latent-variable framework for modeling Chain-of-Thought (CoT) prompting~\cite{hu2024unveilingstatisticalfoundationschainofthought}, enabling in-context learning (ICL) experiments under controlled out-of-distribution conditions. While prior works~\cite{zhang2023doesincontextlearninglearn,xie2022explanationincontextlearningimplicit} suggest that ICL performance is tightly linked to pretraining priors, and that CoT can improve generalization via structured reasoning~\cite{xiao2024theory,cho2024positioncouplingimprovinglength}, our findings indicate that neither approach alone ensures robustness under latent disjointness. These explorations further underscore the need for principled generalization guarantees beyond empirical heuristics~\cite{min2022rethinkingroledemonstrationsmakes,wei2022chain,zhan2022evaluating}. 

Detailed literature survey and the comparison to our work can be found in Sec.~\ref{sec:related_works}. Our contributions are summarized as follows:
\begin{itemize}
  \item We provide the first theoretical upper bound on MSE prediction error of Transformer models under semantic distribution shift, using Wasserstein-1 geometry and Gevrey-class regularity.
  \item We design controlled experimental setups and show the empirical error aligns with our theoretical predictions.
  \item We adapt a latent-variable CoT framework to study ICL behavior under structured OOD scenarios, offering additional evidence of the limitations of prompt-based alignment in semantically disjoint regimes.
\end{itemize}

\section{Problem Formulation}
\label{sec:problem}

In this section, we formalize the tasks and frameworks used in our experiments. We begin with a simple deterministic task designed to isolate progressive reasoning in multiple steps, described in Sec.~\ref{sec:multi_calc}. We then introduce a latent-variable framework for modeling ICL with Chain-of-Thought (CoT) prompting, along with structured forms of distribution shift—specifically, latent permutations and scalings—detailed in Sec.~\ref{sec:cot_framework}. While the former task provides a minimal setting to study step-wise reasoning, the latter captures more realistic ICL dynamics where each prompt is sampled from a distribution over latent tasks. These structured shifts allow us to investigate how LLMs generalize under out-of-distribution (OOD) conditions, and provide the empirical foundation for our theoretical analysis of Wasserstein-type perturbations. The effects of these shifts are visualized in Figures~\ref{fig:result_tilde} and~\ref{fig:result_bar}, which demonstrate how generalization degrades under increasing task-space perturbation.

\subsection{Multiple Steps Calculation Task}
\label{sec:multi_calc}

We begin with a deterministic arithmetic task designed to test multi-step numerical reasoning in a minimal setting. Given a sequence of four real-valued inputs, the model must first compute their average, and then compute the square of that average. Unlike latent-variable CoT setups, this task is fully deterministic and fully observable: compute \((y_0, y_1)\) from \(\mathbf{x} = (x_0, x_1, x_2, x_3)\). We describe the notation, data generation, and training/inference setup below.

\subsubsection{Notation}
\begin{itemize}
  \item $\mathrm{frac}(x)$ : Fractional parts of $x$.
  \item \(\mathbf{x} = (x_0, x_1, x_2, x_3)\), where inputs are sampled from a training or testing support with non-overlapping fractional parts.
  \item Inputs are encoded as space-separated strings:
    \[
      \texttt{input}: \texttt{"}x_0\;x_1\;x_2\;x_3\texttt{"}.
    \]
  \item The two-step target outputs are:
    \[
      y_0 = f_0(x_0, x_1, x_2, x_3),\quad
      y_1 = f_1(x_0, x_1, x_2, x_3,y_0).
    \]
  \item The complete output is:
    \[
      \texttt{output}: \texttt{"}y_0\;y_1\texttt{"}.
    \]
\end{itemize}

\subsubsection{Training Process}
\label{sec:multistep_calc_training}
Training is framed as a progressive, multi-step regression task with teacher forcing: each predicted value is conditioned on prior ground-truth values.

\paragraph{Step 1: Predict \(y_0\)}  
\[
  \texttt{input}_1 = \texttt{"}x_0\;x_1\;x_2\;x_3\texttt{"}\Longrightarrow\; \hat y_0.
\]

\paragraph{Step 2: Predict \(y_1\)}  
\[
  \texttt{input}_2 = \texttt{"}x_0\;x_1\;x_2\;x_3\;y_0\texttt{"}\Longrightarrow\; \hat y_1.
\]

The total training loss is the sum of two mean squared errors:
\[
  \mathcal{L}_{\mathrm{train}} = \mathrm{MSE}(\hat{y}_0, y_0) + \mathrm{MSE}(\hat{y}_1, y_1).
\]

\subsubsection{Inference Prompting}\label{sec:multistep_calc_inference}
At inference time, the model receives an input string and performs autoregressive prediction: each predicted value is fed back into the next stage's input.

\[
  \texttt{prompt} = \texttt{"}x_0\;x_1\;x_2\;x_3\texttt{"}.
\]

The model generates:
\[
  \hat z_0,\;\hat z_1.
\]
with \(\hat z_0 \approx y_0\), and \(\hat z_1 \approx y_1\). For evaluation, we compute mean squared error on the final prediction:
\[
  \mathcal{L}_{\mathrm{eval}} = \mathrm{MSE}(\hat z_1, y_1).
\]

\subsection{ICL with CoT Framework and Distribution Shifts}
\label{sec:cot_framework}

We adopt a generative framework for Chain‐of‐Thought (CoT) demonstrations following~\cite{hu2024unveilingstatisticalfoundationschainofthought}, in which each reasoning sequence is conditioned on an unobserved task‐level latent variable \(\theta \in \Theta\).  Below we describe the model, its pretraining and prompting, the in‐context OOD demonstration scenarios, and the distance‐based interpretation of our shifts.

\subsubsection{Notation}
\begin{itemize}
  \item \(\theta\): A latent variable that represents the underlying task or concept. 
  \item \(\Theta\): The domain (or space) of all possible latent tasks. We assume that \(\theta\) is drawn from a prior distribution \(\pi(\theta)\) on \(\Theta\).
  \item \(n\): The number of demonstration examples provided in the prompt.
  \item \(H\): The length (number of steps) of the chain-of-thought associated with a task in \(\Theta\). In each demonstration \(s^i\), the chain consists of one input, \(H-1\) intermediate steps, and one final output.
  \item  \(\Upsilon_n\): The set of demonstration examples provided in the prompt. When there are \(n\) demonstrations, we write
  \[
    \Upsilon_n = \{ s^1, s^2, \dots, s^n \}.
  \]
  \item Each demonstration \(s^i\) is a full chain-of-thought example sampled conditionally on the latent task, and is structured as a sequence of tokens (or steps)
  \[
    s^i = \bigl(z_0^i, z_1^i, \dots, z_H^i\bigr),
  \]
  where
\(z_0^i\) represents the input for the \(i\)-th demonstration,
     \(z_1^i, z_2^i, \dots, z_{H-1}^i\) denote the intermediate reasoning steps, and
     \(z_H^i\) denotes the final answer for the \(i\)th demonstration.

\item 
 For a test instance, the input is similarly denoted by \(z_0^{\text{test}}\). 
 The LLM is prompted with the full prompt
  \[
    \operatorname{prompt}_{\mathrm{CoT}}(n) = \bigl( \Upsilon_n, \; z_0^{\text{test}} \bigr),
  \]
  and it then autoregressively generates a chain-of-thought \(z_1^{\text{test}}, \dots, z_H^{\text{test}}\) that leads to the final answer.
\end{itemize}

\subsubsection{Generative Process}
Given the above notations, we first specify the task-specific distribution \( \mathbb{P}(\cdot \mid \theta) \), which generates the data for the CoT prompts.  
For each \( \theta \in \Theta \),  $\theta$ consists of the combination for a sequence of latent variables $\theta=(\vartheta_0,\vartheta_1,\vartheta_2,...,\vartheta_H)$, and we generate each reasoning sequence \(s_{i} = z_{0:H}^i \) according to the following stochastic dynamic model:
\begin{equation}\label{eq:dataset_generation}
\begin{aligned}
\mathbb{P}(s^i \mid \theta^* = \theta): 
& z_0^i = f_{\vartheta_0}(\zeta^i), ~ z_h^i = F_{\vartheta_h}(z_{h-1}^i), \\
&  \forall 1 \leq h \leq H,
\end{aligned}
\end{equation}
where \( \{\zeta^i\}_{i \in [n]} \) are i.i.d. noise variables, while \( f_\theta \) and \( F_\theta \) are functions parameterized by \( \theta \in \Theta \).
This framework also applies to the test sequence \( z_{0:H}^{\text{test}} \), defining the target distribution that the LLM learns from during prompting. 
\subsubsection{Pretraining of the LLM:}  
  The LLM is first pretrained via maximum likelihood estimation on a large dataset. 
Specifically, the LLM is pretrained using $\mathscr{L}$ instances. Each instance is independently generated according to the model described in Eq.(\ref{eq:dataset_generation}), parameterized by a task-specific concept $\theta$, drawn independently from distribution $\pi$ for each $\ell \in [\mathscr{L}]$. Each instance $\ell \in [\mathscr{L}]$ comprises $n$ demonstrations $\{s^{i, \ell}\}_{t=1}^T$, independently sampled from the same model (Eq.(\ref{eq:dataset_generation})) under the fixed task latent variable $\theta$, with each example represented as $s^{i, \ell} = (z_0^{i, \ell}, \dots, z_H^{i, \ell})$. Therefore, the entire training dataset consists of $\mathscr{L} n$ examples spanning diverse latent variables.
Let $\{\mathbb{P}_\rho \mid \rho \in \mathcal{P}_{\text{LLM}}\}$ denote the conditional probability distributions generated by the LLM parameterized by $\rho$, with parameter space $\mathcal{P}_{\text{LLM}}$. Pretraining the autoregressive LLM thus corresponds to obtaining the maximum likelihood estimator (MLE) $\widehat{\rho}$, formally expressed as:
$$
\widehat{\rho} = \arg\min_{\rho \in \mathcal{P}_{\mathrm{LLM}}}  \frac{-\sum_{\ell,i,h} \log \mathbb{P}_\rho\left(z_h^{i, \ell} \mid \Upsilon_{i-1, \ell}, \{z_j^{i, \ell}\}_{j=0}^{h-1}\right) }{\mathscr{L} n (H+1)}, 
$$
where $\Upsilon_{i, \ell} = \{s^{k, \ell}\}_{k \in [i]}$ represents the first $i$ examples from the $\ell$-th instance. 

\subsubsection{In‐Context OOD Demonstrations}
\label{sec:CoT_mixed}
We have so far described CoT prompting when all demonstrations share the same latent variable \(\theta\in\Theta\). In practice, a user may instead provide examples from an out‑of‑distribution latent variable \(\theta'\in\Theta'\). We also assume the test input \(z_0^{\mathrm{test}}\) is drawn from this OOD latent variable, so that the true conditional distribution of the answer is
\[
  Q\bigl(y^{\mathrm{test}}\mid z_0^{\mathrm{test}}\bigr)
  = \mathbb{P}\bigl(y^{\mathrm{test}}\mid z_0^{\mathrm{test}},\,\theta'\bigr).
\]
The CoT prompt then becomes
\begin{equation}\label{eq:prompt_cot}
  \operatorname{prompt}_{\mathrm{CoT}}(n)
  = \bigl(\Upsilon_{n}^{\theta'},\,z_0^{\mathrm{test}}\bigr),
  \quad
  \Upsilon_{n}^{\theta'} = \{s_1^{\theta'}, \dots, s_n^{\theta'}\},
\end{equation}
where each demonstration \(s=(z_0,z_1,\dots,z_H)\) is sampled from the multi‑step latent‑variable model conditioned on \(\theta'\), using Eq.~(\ref{eq:dataset_generation}). We focus on the case \(\Theta'\cap\Theta=\emptyset\), detailed below.

\paragraph{$\widetilde{\Theta}$ -- Permuted Combinations of Latent Variables:}
Assume each latent variable $\vartheta_i$ can take $M$ distinct values, written as $\vartheta_i^{(m)}$ for $m\in[M]$.  The full latent variable space is
\[
  \Theta^{\star}
  = \bigl\{(\vartheta_0^{(m_0)}, \vartheta_1^{(m_1)}, \dots, \vartheta_H^{(m_H)})
      \mid m_h\in[M],\;h\in[H]\bigr\}.
\]
We then choose two disjoint subsets $\Theta,\widetilde{\Theta}\subset\Theta^{\star}$.  Define
\[
  \operatorname{FlattenSet}(\vartheta_0,\dots,\vartheta_H)
  = \{(\vartheta_h,h)\mid h=0,1,\dots,H\}.
\]
We require
\begin{equation}\label{eq:theta_tilde}
  \bigcup_{\theta\in\Theta}\operatorname{FlattenSet}(\theta)
  =\;
  \bigcup_{\widetilde{\theta}\in\widetilde{\Theta}}\operatorname{FlattenSet}(\widetilde{\theta}),
\end{equation}
so that $\Theta$ and $\widetilde{\Theta}$ contain exactly the same set of $(\vartheta_h,h)$ pairs, merely combined in different ways.

\paragraph{$\bar{\Theta}$ -- Scaled Variants of Latent Variables:}
Here we introduce a second OOD scenario by scaling each latent variable in $\Theta$.  Formally, we define
\begin{equation}\label{eq:theta_bar}
\begin{aligned}
  \bar{\Theta}
  = \bigl\{\bar{\theta}
    \mid 
    &\bar{\theta}=(p\,\vartheta_0,\;p\,\vartheta_1,\;\dots,\;p\,\vartheta_H),
    \;\\
    &\theta=(\vartheta_0,\vartheta_1,\dots,\vartheta_H)\in\Theta
  \bigr\}.
\end{aligned}
\end{equation}
Here, \(p \neq 1\) is a non-unit constant (either \(p > 1\) or \(p < 1\)) that scales all latent variables in \(\Theta\).  To ensure \(\bar{\Theta}\cap\Theta=\emptyset\), we let 
\[
\Phi = \{\vartheta \mid \vartheta \text{ appears in some } \theta\in\Theta\}
\]
be the set of all latent values occurring in \(\Theta\), and $p$ should satisfies
\begin{equation}\label{eq:p_constraint}
    p\vartheta \notin \Phi \quad \text{for all $\vartheta$ in $\Theta$}. 
\end{equation}

\subsubsection{Distance‐Based Shift Interpretation}
Although defined at the latent level, these shifts induce quantifiable perturbations in input space.  In the scaling case with \(p=1\pm\delta\), generated inputs \(x=f_{\vartheta_0}(\zeta)\) diverge from in‐distribution inputs \(x'\), with \(\|x-x'\|\) growing in \(\delta\).  Empirically (Figs.~\ref{fig:result_tilde},~\ref{fig:result_bar}), test‐time loss increases sub‐exponentially in the shift size, matching the theoretical Wasserstein‐1 bound of Sec.~\ref{sec:theory}.

\paragraph{Design Summary.}
Our experiments adopt a deliberately minimal prompting setup: two OOD demonstrations are provided alongside one test input, forming a controlled two-shot prompt. This stripped-down configuration enables precise measurement of the model’s sensitivity to structured distributional shifts, while aligning cleanly with the assumptions of our theoretical analysis. Although our theoretical results make no assumptions about CoT or ICL prompting, this setup provides a concrete and interpretable empirical testbed for analyzing generalization under distributional drift.

We revisit this connection formally in Sec.~\ref{sec:theory}, where we show that Gevrey-regular LLMs admit provable robustness bounds under Wasserstein-type perturbations.

\section{Experiments}
\label{sec:experiments}

To empirically investigate how structured distribution shifts affect generalization in large language models (LLMs), we conduct a series of controlled experiments under the settings in Sec.~\ref{sec:problem}. Our experimental suite begins with a pedagogically motivated mean calculation task, designed to isolate and study progressive reasoning behavior in a minimal setting (Sec.~\ref{sec:exp_sqpred}--\ref{sec:exp_sq_results}). We then extend this analysis to more complex scenarios involving latent-variable sequence generation (Sec.~\ref{sec:exp_icl_with_cot}--\ref{sec:distance_shift}), where we evaluate generalization performance under two distinct forms of out-of-distribution (OOD) shifts: latent permutations (\(\widetilde{\Theta}\)) and latent scalings (\(\bar{\Theta}\)). These perturbations induce systematic changes in the input distribution, which, as discussed in Sec.~\ref{sec:theory}, correspond to Wasserstein-type shifts in the functional input space. Our goal is to determine whether these structured OOD settings yield test-time behaviors that reflect the theoretical predictions---particularly in the scaling and composition of loss across inference steps.

\subsection{Mean Square Calculation Task Setting}
\label{sec:exp_sqpred}

Each sample consists of four real-valued inputs:
\[
    \mathbf{x} = (x_0, x_1, x_2, x_3), \quad x_i \in [0, n).
\]
For the training set, we constrain \(\mathrm{frac}(x_i) < 0.5\); for the testing set, we require \(\mathrm{frac}(x_i) \geq 0.5\), ensuring a strict distributional shift between training and evaluation.
The input is represented as a space-separated string:
\[
    \texttt{input}: \texttt{"}x_0\;x_1\;x_2\;x_3\texttt{"}.
\]
The corresponding outputs are:
\[
    y_0 = \frac{x_0 + x_1 + x_2 + x_3}{4},\quad
    y_1 = y_0^2,
\]
and thus
\[
    \texttt{output}: \texttt{"}y_0\;y_1\texttt{"}.
\]
The training and inference procedure follows the context in Sec.~\ref{sec:multistep_calc_training} and ~\ref{sec:multistep_calc_inference}.









\subsection{Experiments on Mean Square Calculation}
\label{sec:exp_sq_results}

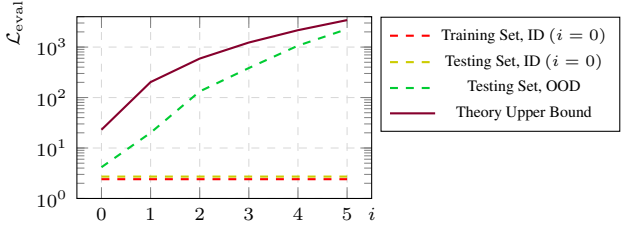
\begin{figure}[h]
  \centering
  \begin{tikzpicture}
\begin{axis}[
    width=5.5cm,
    height=4cm,
    legend pos=outer north east,
    grid=major,
    grid style={dashed,gray!30},
    xlabel={$i$},
    ylabel={$\mathcal{L}_{\mathrm{eval}}$},
    ymode=log,
    ymin=1,
    ymax=4000,
    xtick={0,1,2,3,4,5},
    font=\scriptsize,
    legend style={font=\tiny},
        xlabel style={
            at={(current axis.south east)}, 
            anchor=north east,              
            yshift=15pt,                   
            xshift=5pt                      
        },
        ylabel style={
            at={(current axis.north west)}, 
            anchor=north east,              
            yshift=-5pt,                   
            xshift=10pt                      
        },
]
\addplot[pc1, thick, dashed] table[row sep=\\] {
x y \\
0  2.409582 \\
1  2.409582 \\
2  2.409582 \\
3  2.409582 \\
4  2.409582 \\
5  2.409582 \\
};
\addlegendentry{Training Set, ID $(i=0)$}
\addplot[pc3, thick, dashed] table[row sep=\\] {
x y \\
0  2.713915 \\
1  2.713915 \\
2  2.713915 \\
3  2.713915 \\
4  2.713915 \\
5  2.713915 \\
};
\addlegendentry{Testing Set, ID $(i=0)$}
\addplot[pc5, thick, dashed] table[row sep=\\] {
x y \\
0 4.160364 \\
1 20.032300 \\
2 132.287617 \\
3 384.416731 \\
4 1066.773096 \\
5 2272.587585 \\
};
\addlegendentry{Testing Set, OOD}
\addplot[color=purple!70!black, thick] table[row sep=\\]{
x y \\
0 23.00653306507886 \\
1 202.43979474859285 \\
2 591.8226011387976 \\
3 1230.784697089121 \\
4 2160.9018076733923 \\
5 3424.4561741058146 \\
};
\addlegendentry{Theory Upper Bound}


\end{axis}
\end{tikzpicture}
  \caption{Loss on training and testing sets  for the mean square task.}
  \label{fig:loss_mean_squared}
\end{figure}

We conduct experiments on the mean square prediction task described in Sec.~\ref{sec:exp_sqpred}. The model is trained in a multi-step manner, with explicit supervision at each stage. At evaluation time, predictions are performed autoregressively, with the output of one step used as input to the next.
To simulate distribution shift, we construct six OOD testing sets \(\bar{\Xi}_i\) for \(i = 0, 1, \dots, 5\). Each \(\bar{\Xi}_i\) is sampled from the interval \([i, i+10)\) and constrained to values with \(\mathrm{frac}(x) \geq 0.5\) , while in-distribution (ID) training set and testing set remains $i=0$ and $\mathrm{frac}(x) < 0.5$. This produces a sequence of testing sets with increasing dissimilarity to the training distribution.
We report the mean squared error (MSE) of the final prediction \(\hat y_1\) on each testing set. Results are shown in Figure~\ref{fig:loss_mean_squared}, with performance plotted on a log scale.

\paragraph{Findings.}
As illustrated in Figure~\ref{fig:loss_mean_squared}, model performance degrades as the OOD index \(i\) increases, consistent with the increasing input shift. This confirms the expected sensitivity of autoregressive multi-step reasoning to compounding distributional drift.

\begin{itemize}
  \item Despite ground-truth guidance during training, errors accumulate during inference due to the mismatch between training (teacher-forced) and test (autoregressive) dynamics.
  \item The exponential nature of the squared prediction amplifies small initial errors in the mean, which are further exacerbated across shifts.
\end{itemize}

These results highlight the challenge of structured multi-step inference under input shift, even in simple numerical reasoning tasks.

\subsection{ICL with CoT Settings}\label{sec:exp_icl_with_cot}
We begin with a concrete example to illustrate our setup. Suppose \(H=2\), \(\vartheta_h\in\{-2,-1,1,2\}\) for all \(h\in[H]\).  The noise \(\zeta\) is drawn from \(\mathrm{Uniform}([-0.5,0.5])\) for all \(h\).  The generation functions are
\[
\begin{aligned}
f_{\vartheta_0}(\zeta) 
&= \operatorname{LeakyRelu}(\zeta + \vartheta_0), \\
F_{\vartheta_h}(z_{h-1}) 
&= \operatorname{LeakyRelu}(z_{h-1} + \vartheta_h),
\end{aligned}
\]
where
\[
f(x)=
\begin{cases}
  x, & x>0,\\
  0.5x, & x\le0.
\end{cases}
\]
The full set \(\Theta^{\star}\) of latent‐parameter combinations is:
\[
\begin{tabular}{ll}
  \(\theta_0=(-2,-2,-2)\) & \(\theta_1=(-2,-2,-1)\) \\
  \(\theta_2=(-2,-2,1)\)  & \(\theta_3=(-2,-2,2)\)  \\
  \(\theta_4=(-2,-1,-2)\) & \(\theta_5=(-2,-1,-1)\) \\
  \(\cdots\)             & \(\cdots\)             \\
  \(\theta_{60}=(2,2,-2)\) & \(\theta_{61}=(2,2,-1)\) \\
  \(\theta_{62}=(2,2,1)\)  & \(\theta_{63}=(2,2,2)\) . 
\end{tabular}
\]
Some example samples \(s_i\) generated by two choices of \(\theta\) are:
$$
\begin{aligned}
\theta=(-2, 1, 2) \Rightarrow 
s_1&=(-0.976, 0.024, 2.024), \\
s_2&=(-0.892, 0.108, 2.108), \\
s_3&=(-0.949, 0.051, 2.051) \\
\theta=(1, -2, 1) \Rightarrow 
s_1&=(-0.476, -1.238, -0.119), \\
s_2&=(-0.392, -1.196, -0.098), \\
s_3&=(-0.449, -1.224, -0.112). \\
\end{aligned}
$$
For instance, \(s_1=(-0.976,0.024,2.024)\) is obtained by
$$
\begin{aligned}
\zeta&=\operatorname{UniformSample}([-0.5,0.5]) = 0.048&\\
z_0&= f_{\vartheta_0}(0.048) = \operatorname{LeakyRelu}(0.048-2) = -0.976, \\
z_1&= f_{\vartheta_1}(-0.976) =\operatorname{LeakyRelu}(-0.976+1)= 0.024, \\
z_2&=f_{\vartheta_2}(0.024) =\operatorname{LeakyRelu}(0.024+2) = 2.024.
\end{aligned}
$$
To evaluate CoT generalization under OOD samples, we consider both:
\begin{itemize}
  \item \textbf{Permuted Combination (\(\widetilde{\Theta}\)).}  
    Randomly partition \(\Theta^\star\) into training subset \(\Theta\) and its complement \(\widetilde{\Theta}\), varying \(\lvert\widetilde{\Theta}\rvert/\lvert\Theta\rvert\).
  \item \textbf{Scaled Latents (\(\bar{\Theta}\)).}  
    Scale each \(\vartheta_h\in\Theta\) by \(p=1\pm\delta\) with \(\delta\in[0.05,0.5]\), corresponding to the input‐level shifts of Sec.~\ref{sec:distance_shift}.
\end{itemize}

For our experiments, we adopt GPT-2 \cite{radford2019language} as backbone, but replace each token’s embedding with a scalar \(z_h\) for \(0\le h<H\), leaving positional embeddings unchanged.  The model outputs one scalar per step, trained with MSE loss:
\[
\widehat{\rho}
= \arg\min_{\rho \in \mathcal{P}_{\mathrm{LLM}}}
\frac{\sum_{\ell,h}
\Bigl(
  \mathbb{F}_{\rho}\bigl(\Upsilon_{n-1,\ell},\{z_j^{n,\ell}\}_{j=0}^{h-1}\bigr)
  - z_h^{n,\ell}
\Bigr)^{2}}{\mathscr{L}(H+1)}
,
\]
where \(\Upsilon_{n-1,\ell}\) is the prompt of \(n-1\) demonstrations.  After pre-training, we evaluate step-wise test loss:
\[
\mathcal{L}_{\mathrm{test}}(h)
= \frac{1}{\mathscr{L}'}
\sum_{\ell=1}^{\mathscr{L}'}
\Bigl(
  \mathbb{F}_{\rho}(\Upsilon_{n-1,\ell},\{z_j^{n,\ell}\}_{j=0}^{h-1})
  - z_h^{n,\ell}
\Bigr)^2.
\]
We set \(\mathscr{L}=768{,}000\), \(\mathscr{L}'=7{,}680\), and \(n=20\).  Implementation and data are available\footnote{\url{https://anonymous.4open.science/r/Robustness_transformer_ood/}}.

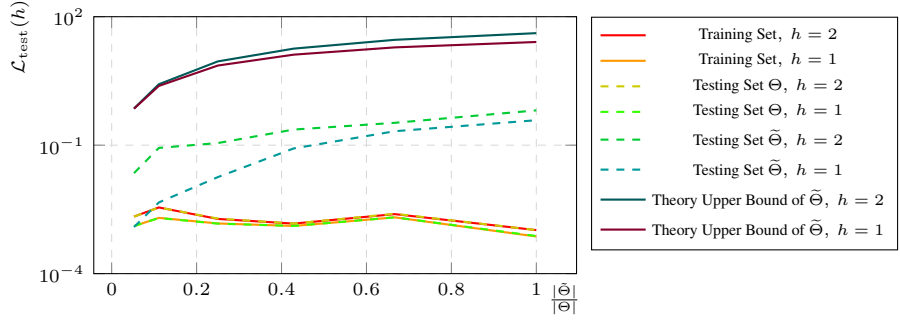
\begin{figure*}[h]
  \centering
  \begin{tikzpicture}
    \begin{axis}[
        width=8cm,
        height=5cm,
        legend pos=outer north east,
        grid=major,
        grid style={dashed,gray!30},
        xlabel={$\frac{|\tilde{\Theta}|}{|\Theta|}$},
        ylabel={$\mathcal{L}_{\mathrm{test}}(h)$},
        title={},
        ymin=0.0001,
        ymax=100,
        ymode=log,
        font=\scriptsize,
        xlabel style={
            at={(current axis.south east)}, 
            anchor=north east,              
            yshift=15pt,                   
            xshift=5pt                      
        },
        ylabel style={
            at={(current axis.north west)}, 
            anchor=north east,              
            yshift=0pt,                   
            xshift=10pt                      
        },
        legend style={
            font=\tiny
        },
        title style={
            font=\normal
        }
    ]
    \addplot[pc1, thick] table[row sep=\\]{
x y \\ 
 0.052631578947368425  0.0021492395986570044 \\ 
 0.11111111111111112  0.0035114395519485696 \\ 
 0.25  0.0018934499938040973 \\ 
 0.4285714285714286  0.0014733631935087033 \\ 
 0.6666666666666667  0.0024533529736800123 \\ 
 1.0  0.0010392274183686824 \\ 
};
\addlegendentry{$\text{Training Set},~h=2$}
\addplot[pc2, thick] table[row sep=\\]{
x y \\ 
 0.052631578947368425  0.0012882057126262225 \\ 
 0.11111111111111112  0.0020059266302268953 \\ 
 0.25  0.0014855536966933867 \\ 
 0.4285714285714286  0.0012980902916751801 \\ 
 0.6666666666666667  0.0020616721056285316 \\ 
 1.0  0.0007374455744866282 \\ 
};
\addlegendentry{$\text{Training Set},~h=1$}
\addplot[pc3, thick, dashed] table[row sep=\\]{
x y \\ 
 0.052631578947368425  0.0021419583383249117 \\ 
 0.11111111111111112  0.003502263489644974 \\ 
 0.25  0.0019296644706628286 \\ 
 0.4285714285714286  0.0014774495619349181 \\ 
 0.6666666666666667  0.002494357293471694 \\ 
 1.0  0.0010387973081378732 \\ 
};
\addlegendentry{$\text{Testing Set} ~\Theta,~h=2$}
\addplot[pc4, thick, dashed] table[row sep=\\]{
x y \\ 
 0.052631578947368425  0.0012876463850261643 \\ 
 0.11111111111111112  0.001992122319643386 \\ 
 0.25  0.0014705436216900123 \\ 
 0.4285714285714286  0.0012992356947506777 \\ 
 0.6666666666666667  0.0020631602819776163 \\ 
 1.0  0.0007555853379017207 \\ 
};
\addlegendentry{$\text{Testing Set} ~\Theta,~h=1$}
\addplot[pc5, thick, dashed] table[row sep=\\]{
x y \\ 
 0.052631578947368425  0.022272099892143158 \\ 
 0.11111111111111112  0.08629889666335656 \\ 
 0.25  0.11198000684380532 \\ 
 0.4285714285714286  0.23143977858126163 \\ 
 0.6666666666666667  0.33255922235548496 \\ 
 1.0  0.6494082234799862 \\ 
};
\addlegendentry{$\text{Testing Set} ~\widetilde{\Theta},~h=2$}
\addplot[pc6, thick, dashed] table[row sep=\\]{
x y \\ 
 0.052631578947368425  0.0012260676972800865 \\ 
 0.11111111111111112  0.004595821531256661 \\ 
 0.25  0.017935230367584154 \\ 
 0.4285714285714286  0.08377979313954712 \\ 
 0.6666666666666667  0.21168140741065145 \\ 
 1.0  0.38255202248692516 \\ 
};
\addlegendentry{$\text{Testing Set} ~\widetilde{\Theta},~h=1$}
\addplot[color=teal!70!black, thick] table[row sep=\\]{
x y \\
0.0526 0.71971 \\
0.1111 2.64485 \\
0.25 9.03930 \\
0.4286 17.96400 \\
0.6667 28.83534 \\
1.0 41.29767 \\
};
\addlegendentry{Theory Upper Bound of $\widetilde{\Theta},~h=2$}

\addplot[color=purple!70!black, thick] table[row sep=\\]{
x y \\
0.0526 0.70951 \\
0.1111 2.41808 \\
0.25 7.23411 \\
0.4286 12.98546 \\
0.6667 19.20406 \\
1.0 25.67554 \\
};
\addlegendentry{Theory Upper Bound of $\widetilde{\Theta},~h=1$}
\end{axis}
 \end{tikzpicture}
  \caption{Test loss on \(\widetilde{\Theta}\) vs.\ \(\Theta\).}
  \label{fig:result_tilde}
\end{figure*}

\begin{figure*}[h]
  \centering
      \begin{tikzpicture}
    \begin{axis}[
        width=8cm,
        height=5cm,
        legend pos=outer north east,
        grid=major,
        grid style={dashed,gray!30},
        xlabel={$\delta$},
        ylabel={$\mathcal{L}_{\mathrm{test}}(h)$},
        title={},
        ymin=0.0001,
        ymax=100,
        ymode=log,
        font=\scriptsize,
        xlabel style={
            at={(current axis.south east)}, 
            anchor=north east,              
            yshift=15pt,                   
            xshift=5pt                      
        },
        ylabel style={
            at={(current axis.north west)}, 
            anchor=north east,              
            yshift=0pt,                   
            xshift=10pt                      
        },
        legend style={
            font=\tiny
        },
        title style={
            font=\normal
        }
    ]
    \addplot[pc1, thick] table[row sep=\\]{
x y \\ 
 0.05  0.002086678788327845 \\ 
 0.1  0.002086678788327845 \\ 
 0.2  0.002086678788327845 \\ 
 0.3  0.002086678788327845 \\ 
 0.4  0.002086678788327845 \\ 
 0.5  0.002086678788327845 \\ 
};
\addlegendentry{$\text{Training Set},~h=2$}
\addplot[pc2, thick] table[row sep=\\]{
x y \\ 
 0.05  0.0014794823352228075 \\ 
 0.1  0.0014794823352228075 \\ 
 0.2  0.0014794823352228075 \\ 
 0.3  0.0014794823352228075 \\ 
 0.4  0.0014794823352228075 \\ 
 0.5  0.0014794823352228075 \\ 
};
\addlegendentry{$\text{Training Set},~h=1$}
\addplot[pc3, thick, dashed] table[row sep=\\]{
x y \\ 
 0.05  0.0020974150770295334 \\ 
 0.1  0.0020974150770295334 \\ 
 0.2  0.0020974150770295334 \\ 
 0.3  0.0020974150770295334 \\ 
 0.4  0.0020974150770295334 \\ 
 0.5  0.0020974150770295334 \\ 
};
\addlegendentry{$\text{Testing Set} ~\Theta,~h=2$}
\addplot[pc4, thick, dashed] table[row sep=\\]{
x y \\ 
 0.05  0.0014780489401649295 \\ 
 0.1  0.0014780489401649295 \\ 
 0.2  0.0014780489401649295 \\ 
 0.3  0.0014780489401649295 \\ 
 0.4  0.0014780489401649295 \\ 
 0.5  0.0014780489401649295 \\ 
};
\addlegendentry{$\text{Testing Set} ~\Theta,~h=1$}
\addplot[pc5, thick, dashed] table[row sep=\\]{
x y \\ 
 0.05  0.011585884973950062 \\ 
 0.1  0.03578070312117537 \\ 
 0.2  0.08097443915903568 \\ 
 0.3  0.06421269970790794 \\ 
 0.4  0.06639543013491979 \\ 
 0.5  0.07520055940064292 \\ 
};
\addlegendentry{$\text{Testing Set} ~\bar{\Theta}, ~p=1-\delta, ~h=2$}
\addplot[pc6, thick, dashed] table[row sep=\\]{
x y \\ 
 0.05  0.006590070845171187 \\ 
 0.1  0.02184207303604732 \\ 
 0.2  0.06511733238585293 \\ 
 0.3  0.05954215376793096 \\ 
 0.4  0.06735860820238788 \\ 
 0.5  0.08713326432431738 \\ 
};
\addlegendentry{$\text{Testing Set} ~\bar{\Theta}, ~p=1-\delta, ~h=1$}
\addplot[pc7, thick, dashed] table[row sep=\\]{
x y \\ 
 0.05  0.003398434005309051 \\ 
 0.1  0.019026296601320308 \\ 
 0.2  0.08247865058171254 \\ 
 0.3  0.20346160093322396 \\ 
 0.4  0.4270299662525456 \\ 
 0.5  0.7926412295550108 \\ 
};
\addlegendentry{$\text{Testing Set} ~\bar{\Theta}, ~p=1+\delta, ~h=2$}
\addplot[pc8, thick, dashed] table[row sep=\\]{
x y \\ 
 0.05  0.0034935981185602333 \\ 
 0.1  0.015243097674101591 \\ 
 0.2  0.06421687120261292 \\ 
 0.3  0.15823278228441875 \\ 
 0.4  0.3003129603341222 \\ 
 0.5  0.5019747421145438 \\ 
};
\addlegendentry{$\text{Testing Set} ~\bar{\Theta}, ~p=1+\delta, ~h=1$}
\addplot[color=teal!70!black, thick] table[row sep=\\]{
x y \\
0.05 0.18126 \\
0.1 0.71026 \\
0.2 2.41844 \\
0.3 4.66047 \\
0.4 7.23411 \\
0.5 10.03082 \\
};
\addlegendentry{Theory Upper Bound of $\bar{\Theta}, h=1$}
\addplot[color=purple!80!black, thick] table[row sep=\\]{
x y \\
0.05 0.18209 \\
0.1 0.72050 \\
0.2 2.64528 \\
0.3 5.47452 \\
0.4 9.03930 \\
0.5 13.22862 \\
};
\addlegendentry{Theory Upper Bound of $\bar{\Theta}, h=2$}
\end{axis}
 \end{tikzpicture}
  \caption{Test loss on \(\bar{\Theta}\) vs.\ \(\Theta\).}
  \label{fig:result_bar}
\end{figure*}
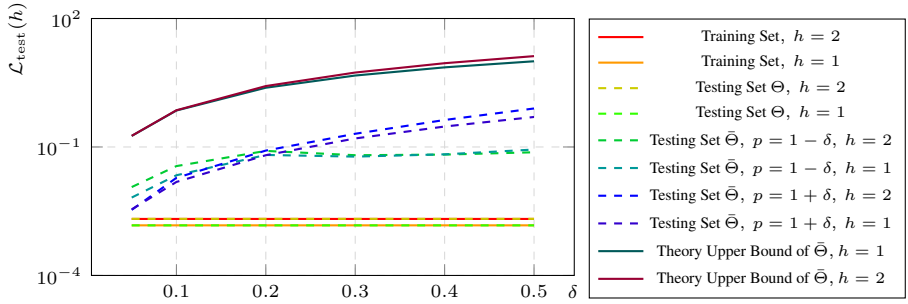

\subsection{Experiments on ICL with CoT}
\label{sec:distance_shift}
In this section, we evaluate one in-distribution scenario and two out-of-distribution scenarios:
\begin{itemize}
  \item \textbf{Testing Set \(\Theta\).}  
    We use the same latent variables \(\theta\in\Theta\) as in the training set, but generate each instance via Eq.~(\ref{eq:dataset_generation}) with a different random seed of \(\zeta\). Hence, the samples follow the same distribution yet are distinct from the training examples. Fig.~\ref{fig:result_tilde} and~\ref{fig:result_bar} show that, despite the new random seed, the LLM achieves losses on these unseen instances that are nearly identical to those on the training set.

  \item \textbf{Testing Set \(\widetilde{\Theta}\).}  
    We obtain \(\widetilde{\Theta}\) by randomly partitioning \(\Theta^\star\) into two disjoint subsets according to Eq.~(\ref{eq:theta_tilde}). Different partitions yield various ratios \(\lvert\widetilde{\Theta}\rvert/\lvert\Theta\rvert\), whose impact is plotted in Fig.~\ref{fig:result_tilde}.  As the ratio \(\lvert\widetilde{\Theta}\rvert/\lvert\Theta\rvert\) decreases, larger portion of combinations of $\vartheta$ are included in the training set, and hence the test loss on \(\widetilde{\Theta}\) also decreases. Although \(\widetilde{\Theta}\cap\Theta=\emptyset\), both sets share the same elements \(\vartheta\) in different combinations, and hence CoT could generalize to this OOD setting.
    Moreover, we find that the test error at \(h=2\) exceeds that at \(h=1\), suggesting that errors accumulate as the CoT unfolds. 
    
  \item \textbf{Testing Set \(\bar{\Theta}\).}  
    We construct \(\bar{\Theta}\) via Eq.~(\ref{eq:theta_bar}), setting \(p=1\pm\delta\) for \(\delta>0\), and both $1+\delta$ and $1-\delta$ should satisfy Eq.~(\ref{eq:p_constraint}). Fig.~\ref{fig:result_bar} illustrates results for several values of \(\delta\). When \(\delta\) is small, the value of latent variables \(\vartheta\) in \(\bar{\Theta}\) remain close to those in \(\Theta\), leading to reduced testing error, despite the fact that \(\bar{\Theta}\cap\Theta=\emptyset\). This further demonstrates CoT’s ability to generalize to these nearby OOD samples. Moreover, the test error is higher for \(p=1+\delta\) than for \(p=1-\delta\), because increasing \(\vartheta\) generates more \(z_h\) values that lie outside the training distribution. 
\end{itemize}

\vspace{0.5em}
\noindent
These experimental findings offer empirical support for our theoretical results: the structured perturbations we introduced in latent task space induce functional shifts in the LLM’s input-output behavior that align with the predicted generalization gap under Wasserstein-1 perturbations. These results qualitatively match the sub-exponential decay behavior formalized in Sec.~\ref{sec:theory}, particularly as a function of the shift size \(\delta\) and the depth \(h\) of CoT inference. We now turn to the formal analysis in Sec.~\ref{sec:theory}.

\section{Theoretical Analysis}
\label{sec:theory}
We begin this section by introducing the notion of Gevrey class functions, which generalize analytic functions by allowing controlled sub-factorial growth in derivatives. In particular, we will focus on the class $G^s$ with $s \geq 1$, and show how this regularity arises naturally from the architecture of transformer-based language models.
The first two subsections are devoted to building this foundation: we introduce Gevrey regularity, and show that recursive auto-regressive language models such as GPT fall into this smoothness class. The third subsection presents our main result, which connects this regularity to generalization under distribution shift.

\subsection{Gevrey Class Functions}

\begin{definition}[Gevrey Class]
Let $f: X \rightarrow Y$ be a smooth function, where $X \subseteq \mathbb{R}^n$ and $Y \subseteq \mathbb{R}^m$. For a fixed $s \geq 1$, the function $f$ is said to belong to the \textbf{Gevrey class} $G^s$ if for every compact subset $K \subset X$, there exist constants $C > 0$ and $R > 0$ such that for all multi-indices $\alpha$ and all $x \in K$, we have:
\[
|\partial^\alpha f(x)| \leq C R^{|\alpha|} (\alpha!)^s.
\]
\end{definition}

The Gevrey class $G^s$ is a generalization of analytic functions. When $s = 1$, $G^1$ coincides with the space of real-analytic functions. For $s > 1$, $G^s$ contains strictly more functions while still enforcing sub-factorial growth in derivatives.
Importantly, for all $s \geq 1$, the Gevrey class $G^s$ is \textbf{closed under composition and pointwise multiplication}. That is, if $f, g \in G^s$, then $f \cdot g \in G^s$, and if $f \in G^s$ and $g \in G^s$ with $g$ mapping into the domain of $f$, then $f \circ g \in G^s$.

\subsection{Recursive Language Models and Gevrey Regularity}

We model an auto-regressive language model as a family of functions:
\[
F_n : X^n \rightarrow X,
\]
where $X = \mathbb{R}^d$ denotes the token embedding space, and $n$ is the sequence length. Each function $F_n$ maps a sequence of $n$ embedded tokens $(x_1, \dots, x_n)$ to the embedding of the next predicted token.
The recursive nature of such models is captured by defining:
\begin{equation}\label{eq:rec_def}
\begin{aligned}
F_{n+1}&(x_1, \dots, x_n, x_{n+1})  \\
&:= F(x_1, \dots, x_n, F_n(x_1, \dots, x_n)),
\end{aligned}
\end{equation}
where $F$ represents the shared transformer architecture, reused across time steps via weight sharing.

\begin{definition}[Recursive Language Model Family]
A \textbf{recursive language model family} is a sequence of functions $\{F_n\}_{n \in \mathbb{N}}$, where each $F_n : X^n \rightarrow X$ satisfies the recurrence relation given in Eq.~\eqref{eq:rec_def}.
\end{definition}
In practice, $F$ is implemented as a stack of $L$ transformer blocks, consisting of:
\begin{itemize}
    \item Multi-head self-attention with softmax activation,
    \item Position-wise feedforward networks with nonlinearities (e.g., GeLU or ReLU),
    \item Layer normalization and residual connections.
\end{itemize}

Although the model output is typically projected to a vocabulary space $Y = \mathbb{R}^V$ before applying softmax, the sampled token is mapped back to $X$ via an embedding lookup. Hence, we may abstract the end-to-end function as $F_n : X^n \rightarrow X$ for theoretical analysis.

\begin{theorem}[Gevrey Smoothness of the Model Family]\label{thm:gevrey_family}
Let $\{F_n\}$ be a recursive language model family as defined above, implemented by a transformer architecture composed of operations including linear maps, softmax, nonlinear activations, normalization, and residual connections. Then for some $s \geq 1$, each $F_n \in G^s(X^n, X)$. That is, the model defines a family of functions with Gevrey regularity of $s$.
\end{theorem}

\begin{proof}[Sketch of Proof]
The transformer architecture is constructed from operations that are either linear (e.g., affine projections), analytic (e.g., softmax), or smooth nonlinearities (e.g., GeLU), each of which lies in $G^s$ for some $s \geq 1$. The Gevrey class $G^s$ is closed under composition and pointwise multiplication.
Given that each $F_n$ is defined recursively through composition of such operations, it follows that $F_n \in G^s(X^n, X)$ for all $n \in \mathbb{N}$.
\end{proof}

A detailed proof is provided in Sec.~\ref{sec:gevery_regularity}.

\subsection{Gevrey Regularity Meets Wasserstein Shift: A New Framework for OOD Generalization}

To our knowledge, this is the first work to analyze the out-of-distribution (OOD) generalization of large language models (LLMs) through the combined lens of Gevrey class smoothness and Wasserstein-1 distance. Existing theoretical approaches to OOD generalization predominantly rely on divergence measures such as KL divergence or domain discrepancy, both of which implicitly require overlapping support between the training and testing distributions. However, in many realistic scenarios—such as domain shift, compositional novelty, or task recombination—the test inputs may lie entirely outside the support of the training distribution, rendering these divergences ill-defined or uninformative.

To address this limitation, we adopt the Wasserstein-1 distance, a metric rooted in optimal transport theory that remains well-defined even when the distributions are disjoint. Unlike KL, which measures pointwise log-likelihood ratios, Wasserstein-1 quantifies the minimal "cost" of transforming one distribution into another under a prescribed geometry. This makes it particularly well-suited for modeling distribution shift in latent semantic spaces, where support overlap cannot be assumed and task structure is geometric in nature.
Instead, we consider a compact-input setting and assume the prediction map $F$ (modeled by the LLM) belongs to a Gevrey class $G^s$ for some $s \geq 1$. We then measure distribution shift using the Wasserstein-1 distance $W_1(P_1, P_2)$ between the training distribution $P_1$ and testing distribution $P_2$. This allows us to quantify how the function $F$ generalizes from $P_1$ to $P_2$, even in the absence of overlapping support.

\begin{theorem}[Main Theorem]\label{thm:main_theorem}
Let $F$ be a Gevrey-class model prediction function and $G$ a Lipschitz target, both defined on a compact domain $K$. Suppose the ID testing error under distribution $P_1$ is at most $\varepsilon$, and $P_2$ is a OOD testing distribution at Wasserstein-1 distance $d$ from $P_1$. Then:
\[
\begin{aligned}
& \mathbb{E}_{x \sim P_2}[\|F(x) - G(x)\|^2] \\
& \quad \leq 6A^2 \cdot \exp\left( -C \cdot d^{-1/(s+1)} \cdot \log\left( \tfrac{1}{d} \right) \right) \\
& \quad  + 3 \varepsilon + 3 L_1^2 \cdot d^2.
\end{aligned}
\]
Here, the first term captures the effect of distribution shift, the second the in-distribution training error, and the third the sensitivity of the target $G$.
\end{theorem}

\textbf{Intuition.} The bound reflects a natural interplay between model smoothness and distributional perturbation. The smoother the model (i.e., smaller $s$), the more resilient it is to small shifts. The Wasserstein distance measures how much mass must be moved to transform $P_1$ into $P_2$, and Gevrey smoothness implies that small input perturbations cause rapidly decaying changes in output—captured by an exponentially decaying modulus of continuity. This leads to a sub-exponential but super-polynomial decay in error as a function of the shift size $d$.

\begin{proof}[Sketch of Proof]
 The proof begins by decomposing the generalization error into a shift term, training error, and a Lipschitz term. The shift term is bounded using the Gevrey modulus of continuity in combination with a tail-based Wasserstein inequality (Lemma~\ref{lem:tail-gevery-bound}). We partition the transport coupling and apply a piecewise upper bound to handle the non-convexity of the modulus function, optimizing the cutoff analytically. The Lipschitz term is controlled using Jensen's inequality. 
\end{proof}

A complete proof and technical details are  in Sec.~\ref{sec:appendix_b}.

\subsection{Quantifying Shift Size via Permutation and Scaling Bounds}
\label{sec:quantifying_shift}

To connect the generalization bound of Main Theorem to the concrete experimental settings of Sec.~\ref{sec:experiments}, we quantify the Wasserstein-1 shift size \(d\) in terms of the experimental parameters \(r = |\widetilde{\Theta}|/|\Theta|\) and \(\delta\), which control the degree of distribution mismatch under permutation and scaling shifts respectively.

In Sec.~\ref{sec:appendix_d}, we establish the following upper bounds:

\begin{itemize}
    \item In the \textbf{mean calculation task}, the test inputs are shifted versions of the training data with support in the intervals \([i, i+10)\), while the training inputs are constrained to \([0, n)\) with fractional parts less than 0.5. This induces a structured shift in the input space. For each integer \(i\), we derive an upper bound on the Wasserstein-1 distance as:
    \[
    d \leq 4(i + 0.5),
    \]
    reflecting the total L1 displacement between matched samples in 4D space.

    \item In the \textbf{permutation setting}, where the testing set \(\widetilde{\Theta}\) is formed by reordering latent components unseen during training, the input shift magnitude is bounded by
    \[
    d \leq D_{\mathrm{max}} \cdot \frac{r}{1 + r},
    \]
    where \(D_{\mathrm{max}} \leq 31.24\) is the maximum Euclidean distance between any two input vectors.
    
    \item In the \textbf{scaling setting}, where each latent component is rescaled as \(\bar{\theta} = (1 \pm \delta) \cdot \theta\), the input shift magnitude is bounded by
    \[
    d \leq 2 \delta \cdot \sqrt{61}.
    \]
\end{itemize}

These results allow us to substitute each bound into the generalization inequality of Theorem~\ref{thm:main_theorem}. In particular, ignoring the additive error terms \(\varepsilon\) and \(d^2\), we obtain the dominant asymptotic behavior of the generalization gap under each setting as:
\[
\log \mathbb{E}_{x \sim P_2}[\|F(x) - G(x)\|^2]
= \widetilde{\mathcal{O}}\left(-d^{-1/(s+1)} \log\left(\tfrac{1}{d}\right)\right),
\]
which, after substituting the bounds on \(d\), becomes:
\begin{align*}
&\log \mathrm{MSE}  \\
&\quad\quad\lesssim
\begin{cases}
-\left(4(i + 0.5)\right)^{-1/(s+1)} \cdot \log\left( \tfrac{1}{4(i + 0.5)} \right), \\
\quad \text{Mean Calculation}, \\
-\left( \tfrac{r}{1+r} \right)^{-1/(s+1)} \cdot \log\left( \tfrac{1+r}{r} \right), \\
\quad \text{Permutation}, \\
-(\delta \sqrt{61})^{-1/(s+1)} \cdot \log\left( \tfrac{1}{\delta} \right), \\ \quad \text{Scaling}.
\end{cases}
\end{align*}

\paragraph{Connection to Empirical Plots.}
Figures~\ref{fig:result_tilde},~\ref{fig:result_bar}, and~\ref{fig:loss_mean_squared} visualize the empirical test losses as functions of the permutation ratio \(r = |\widetilde{\Theta}|/|\Theta|\), the scaling factor \(\delta\), and the index \(i\) in the mean calculation task, respectively. Overlaid on all three figures are theoretical upper bound curves computed using Theorem~\ref{thm:main_theorem} and the Wasserstein-1 estimates from Appendix~\ref{sec:appendix_d}.

The resulting curves exhibit the characteristic sub-exponential but super-polynomial growth predicted by our Gevrey-based generalization framework. Specifically:
\begin{itemize}
    \item In Figure~\ref{fig:loss_mean_squared}, the upper bounds show a consistent upward trend with increasing \(i\), matching the shift magnitude \(4(i + 0.5)\);
    \item In Figure~\ref{fig:result_tilde}, the upper bound curves closely follow the empirical rise in test loss as \(r\) increases, reflecting the dependence on the transport cost term \(\frac{r}{1 + r}\);
    \item In Figure~\ref{fig:result_bar}, the theory captures the pronounced nonlinear growth in \(\log \mathcal{L}_{\mathrm{test}}\) with respect to \(\delta\), in agreement with the Gevrey continuity modulus involving \(-d^{-1/(s+1)} \log(1/d)\).
\end{itemize}

All constants used to generate the theory curves are fixed and shared across experiments, and chosen solely to highlight the asymptotic shape of the upper bound. Together, these results validate that our theoretical predictions meaningfully track the empirical behavior of LLMs under structured latent perturbations.

\section{Limitations}
\label{sec:limitations}

While our theoretical and empirical analyses provide novel insights into generalization under semantic distribution shift, several limitations remain.

\paragraph{Model and Task Simplifications.}  
Our analysis relies on structural simplifications of both model and task. Transformer models are treated as smooth Gevrey-class function approximators, omitting the intricacies of tokenization, sampling, and mixed discrete-continuous behavior inherent in real-world LLMs. Similarly, our experiments focus on synthetic arithmetic and reasoning tasks with known latent structures, which—while enabling clean theoretical alignment—fail to capture the linguistic ambiguity, long-range dependencies, and diverse reasoning strategies found in realistic Chain-of-Thought (CoT) examples. These simplified settings do not reflect the challenges of integrating external knowledge or handling unstructured prompts, which are common in naturalistic applications.

\paragraph{Model Scale and Architecture.}  
Our empirical studies are conducted on GPT-2, a relatively small language model by today’s standards. While this choice allows controlled experimentation and tractable analysis, it limits the generalizability of our conclusions to larger models with billions of parameters and more complex architectural features. Whether similar generalization patterns and robustness properties hold for state-of-the-art models remains an open question.

\paragraph{Restricted OOD Scenarios.}  
We consider two types of structured out-of-distribution shifts—latent permutations and scalings—that only partially reflect the spectrum of real-world distributional drift. These settings allow quantifiable Wasserstein-type comparisons but exclude adversarial, compositional, or concept-drifting scenarios, where LLMs often falter in less predictable ways.

\paragraph{Bounding Tightness.}  
The theoretical upper bounds presented are asymptotic in nature and not tuned for numerical tightness. Constants in the generalization inequality are derived for clarity rather than precision, and the sub-exponential decay terms may overestimate degradation under moderate shifts. These bounds are therefore best viewed as qualitative guides rather than predictive tools.

\section{Conclusion}
\label{sec:conclusion}

In this work, we presented a theoretical and empirical framework for understanding the generalization behavior of large language models (LLMs) under structured semantic distribution shift. By characterizing autoregressive models as members of the Gevrey function class and measuring input perturbations via the Wasserstein-1 distance, we derived provable sub-exponential upper bounds on out-of-distribution (OOD) prediction error. Our experiments on both deterministic arithmetic tasks and Chain-of-Thought (CoT) prompting scenarios revealed that test-time degradation aligns closely with these theoretical predictions, highlighting the utility of smoothness-based approaches to reasoning about robustness.
Importantly, our findings suggest that generalization under latent disjointness is not merely a question of pretraining scale or prompt design, but is fundamentally tied to the structure of the function class realized by the model. While current results are based on simplified tasks and moderate-scale models, they nonetheless provide a blueprint for more principled analyses of LLM behavior in the presence of semantic novelty.

\paragraph{Future Work.} Several avenues merit further investigation. First, extending our framework to modern LLMs with architectural variations (e.g., mixture-of-experts, retrieval-augmented models) could reveal how functional smoothness properties evolve with scale. Second, real-world tasks often involve unstructured natural language, ambiguous reasoning steps, and noisy or missing latent variables. Modeling and bounding generalization in these settings will require richer latent-variable formalisms and new theoretical tools. Finally, our analysis opens the door to designing training curricula and prompts that explicitly optimize geometric robustness metrics—an approach that could move beyond today's empirical prompt tuning and toward generalization-aware inference strategies.

\appendix




\section*{Impact Statement}
This paper presents work whose goal is to advance the field of 
Machine Learning. There are many potential societal consequences 
of our work, none which we feel must be specifically highlighted here.

\bibliography{References}
\bibliographystyle{icml2025}
\clearpage
\section{Gevrey Regularity of Transformer-Based Models}\label{sec:gevery_regularity}
\renewcommand{\thedefinition}{A.\arabic{definition}}
\setcounter{definition}{0}
\renewcommand{\theproposition}{A.\arabic{proposition}}
\setcounter{proposition}{0}

\subsection{Analytic Functions and the Gevrey Hierarchy}

We begin by reviewing the definition of real analytic functions and Gevrey classes, and establishing the foundational relationship between them.

\begin{definition}[Real Analytic Function]
Let $f : U \subseteq \mathbb{R}^n \rightarrow \mathbb{R}^m$ be a smooth function. We say that $f$ is \textbf{real analytic} at a point $x_0 \in U$ if there exists a neighborhood $V \subseteq U$ of $x_0$ such that $f$ admits a convergent Taylor expansion:
\[
f(x) = \sum_{|\alpha|=0}^{\infty} \frac{1}{\alpha!} \partial^\alpha f(x_0) (x - x_0)^\alpha, \quad \forall x \in V.
\]
In other words, the Taylor series of $f$ centered at $x_0$ converges to $f$ in some neighborhood of $x_0$.
\end{definition}

\begin{definition}[Gevrey Class $G^s$]
Let $f : U \subseteq \mathbb{R}^n \rightarrow \mathbb{R}^m$ be a smooth function and $s \geq 1$. We say that $f \in G^s(U)$ if for every compact set $K \subset U$, there exist constants $C, R > 0$ such that for all multi-indices $\alpha \in \mathbb{N}^n$ and all $x \in K$,
\[
|\partial^\alpha f(x)| \leq C R^{|\alpha|} (\alpha!)^s.
\]
Here, the norm $|\cdot|$ can be interpreted componentwise for vector-valued functions: $f \in G^s$ if and only if all coordinate functions $f_i$ belong to $G^s$.
\end{definition}

\begin{proposition}
Let $f : U \subseteq \mathbb{R}^n \rightarrow \mathbb{R}^m$ be real analytic. Then $f \in G^1(U)$.

Moreover, for any compact set $K \subset U$, there exist constants $C, R > 0$ such that for all multi-indices $\alpha \in \mathbb{N}^n$ and $x \in K$, we have
\[
|\partial^\alpha f(x)| \leq C R^{|\alpha|} \alpha!.
\]
\end{proposition}

\begin{proof}
Since $f$ is real analytic, for every $x_0 \in K$ there exists a neighborhood $V_{x_0} \subseteq U$ on which the Taylor series of $f$ converges to $f$:
\[
f(x) = \sum_{\alpha \in \mathbb{N}^n} \frac{1}{\alpha!} \partial^\alpha f(x_0) (x - x_0)^\alpha.
\]
By the theory of analytic functions, this convergence implies the existence of constants $M_{x_0} > 0$ and $\rho_{x_0} > 0$ such that:
\[
|\partial^\alpha f(x_0)| \leq M_{x_0} \rho_{x_0}^{-|\alpha|} \alpha!,
\]
for all multi-indices $\alpha$. This inequality is uniform on some neighborhood of $x_0$.

Now, since $K$ is compact, it can be covered by finitely many such neighborhoods $V_{x_1}, \dots, V_{x_N}$. Let
\[
C := \max_{1 \leq i \leq N} M_{x_i}, \quad R := \max_{1 \leq i \leq N} \rho_{x_i}^{-1}.
\]
Then, for all $x \in K$ and all multi-indices $\alpha$,
\[
|\partial^\alpha f(x)| \leq C R^{|\alpha|} \alpha!,
\]
as required. Hence $f \in G^1(K)$, and since $K$ was arbitrary, $f \in G^1(U)$.
\end{proof}

\paragraph{Why Analyticity Matters.}

Many functions used in deep learning architectures—such as $\tanh$, $\exp$, $\log(1 + e^x)$ (Softplus), and GeLU—are analytic on all of $\mathbb{R}$ and thus belong to $G^1$. Their derivatives grow at most factorially, and their Taylor expansions converge everywhere or on sufficiently large domains. Thus, Gevrey classes (and in particular $G^1$) naturally include the vast majority of activation and normalization functions used in practice.

\paragraph{Gevrey Classes Beyond Analyticity.}

The Gevrey hierarchy \( G^s \) for \( s > 1 \) generalizes analyticity to ultra-smooth functions with slower convergence properties. For example, compactly supported smooth functions (e.g., bump functions) are not analytic but can belong to \( G^s \) for some \( s > 1 \). This provides a flexible framework to analyze model regularity without requiring global analyticity.

\subsection{Gevrey Regularity of Transformer Operations}

Before examining the specific components of transformer architectures, we first formalize the key closure properties of the Gevrey class that will support our analysis.

\begin{proposition}[Closure Properties of Gevrey Classes with Explicit Constants and Orders]\label{prop:gevrey_closure_explicit_full}
Let $U \subseteq \mathbb{R}^n$ be open, and let $f, g$ be smooth functions defined on $U$ or appropriate codomains. The following operations preserve Gevrey regularity with explicit control on constants and Gevrey order:
\begin{enumerate}
    \item \textbf{Addition:} If $f \in G^{s_f}(U)$ with constants $(C_f, R_f)$ and $g \in G^{s_g}(U)$ with $(C_g, R_g)$, then \( f + g \in G^{s_h}(U) \) with $(C_h,R_h)$, where
    \[
\begin{aligned}
C_h &= C_f + C_g, \\
R_h &= \max(R_f, R_g), \\
s_h &= \max(s_f, s_g).
\end{aligned}
\]
    
    \item \textbf{Multiplication:} If $f, g$ are scalar-valued and \( f \in G^{s_f}(U) \), \( g \in G^{s_g}(U) \), then \( fg \in G^{s_h}(U) \) with $(C_h,R_h)$, where
    \[
\begin{aligned}
C_h &= C_f C_g 2^{s_h}, \\
R_h &= R_f + R_g, \\
s_h &= \max(s_f, s_g).
\end{aligned}
\]

    \item \textbf{Cartesian Product:} If each component \( f_i \in G^{s_i}(U) \) with constants \( (C_i, R_i) \), then \( f = (f_1, \dots, f_m) \in G^{s_h}(U, \mathbb{R}^m) \), with
    \[
\begin{aligned}
C_h &= \max_i C_i, \\
R_h &= \max_i R_i, \\
s_h &= \max_i s_i.
\end{aligned}
\]

    \item \textbf{Composition:} Let \( f : V \to \mathbb{R} \in G^{s_f}(V) \) with constants \( C_f, R_f \), and \( g : U \to V \subseteq \mathbb{R}^d \in G^{s_g}(U) \) with constants \( C_g, R_g \), and assume \( g(U) \subset V \). Then:
    \[
f \circ g \in G^{s_h}(U), \quad \text{where}
\]
\[
\begin{aligned}
C_h &= C_f \cdot \exp(C_g R_g), \\
R_h &= R_f \cdot (C_g R_g)^{s_f}, \\
s_h &= s_f s_g.
\end{aligned}
\]
\end{enumerate}
\end{proposition}

\begin{proof}
We treat each operation separately.

\textbf{Addition:} The derivatives of \( h = f + g \) satisfy
\begin{align*}
|\partial^\alpha h(x)| 
&\leq |\partial^\alpha f(x)| + |\partial^\alpha g(x)| \\
&\leq C_f R_f^{|\alpha|} (\alpha!)^{s_f} 
    + C_g R_g^{|\alpha|} (\alpha!)^{s_g}
\end{align*}
Using \( s_h = \max(s_f, s_g) \), we have 
\( (\alpha!)^{s_f} \leq (\alpha!)^{s_h} \) 
and similarly \( (\alpha!)^{s_g} \leq (\alpha!)^{s_h} \). 

Also, since \( R_f^{|\alpha|} \leq R_h^{|\alpha|} \) 
and \( R_g^{|\alpha|} \leq R_h^{|\alpha|} \) 
if \( R_h = \max(R_f, R_g) \), we obtain:
\[
|\partial^\alpha h(x)| \leq (C_f + C_g) R_h^{|\alpha|} (\alpha!)^{s_h}.
\]

\textbf{Multiplication:} For \( h = fg \), use the multivariate Leibniz rule:
\[
\partial^\alpha(fg)(x) = \sum_{\beta \leq \alpha} \binom{\alpha}{\beta} \partial^\beta f(x) \cdot \partial^{\alpha - \beta} g(x).
\]
Using \( |\partial^\beta f(x)| \leq C_f R_f^{|\beta|} (\beta!)^{s_f} \), 
and likewise for \( g \), 

together with the bound \( \binom{\alpha}{\beta} \leq 2^{|\alpha|} \), 

we have:

\[
|\partial^\alpha(fg)(x)| \leq C_f C_g 2^{|\alpha|} \sum_{\beta \leq \alpha} R_f^{|\beta|} R_g^{|\alpha - \beta|} (\beta!)^{s_f} ((\alpha - \beta)!)^{s_g}.
\]
Apply \( \beta!^{s_f} (\alpha - \beta)!^{s_g} \leq \alpha!^{s_h} \), 
with \( s_h = \max(s_f, s_g) \). 

Also, note that 
\( R_f^{|\beta|} R_g^{|\alpha - \beta|} \leq (R_f + R_g)^{|\alpha|} \), 

so

\[
|\partial^\alpha(fg)(x)| \leq C_f C_g 2^{|\alpha|} (R_f + R_g)^{|\alpha|} (\alpha!)^{s_h},
\]
which gives
\[
\begin{aligned}
C_h &= C_f C_g 2^{s_h}, \\
R_h &= R_f + R_g.
\end{aligned}
\]

\textbf{Cartesian Product:} Suppose \( f = (f_1, \dots, f_m) \), each \( f_i \in G^{s_i} \) with \( |\partial^\alpha f_i(x)| \leq C_i R_i^{|\alpha|} (\alpha!)^{s_i} \). Then let:
\[
\begin{aligned}
C_h &= \max_i C_i, \\
R_h &= \max_i R_i, \\
s_h &= \max_i s_i.
\end{aligned}
\]
Then for any \( x \in U \), we bound:
\[
\|\partial^\alpha f(x)\|^2 = \sum_i |\partial^\alpha f_i(x)|^2 \leq m \cdot C_h^2 R_h^{2|\alpha|} (\alpha!)^{2s_h},
\]
so
\[
\|\partial^\alpha f(x)\| \leq \sqrt{m} C_h R_h^{|\alpha|} (\alpha!)^{s_h}.
\]

\textbf{Composition:} Use the multivariate Faà di Bruno formula. Let \( h(x) = f(g(x)) \), where \( f \in G^{s_f}, g \in G^{s_g} \). The formula gives:
\[
\partial^\alpha h(x) = \sum_{\pi} \partial^\beta f(g(x)) \cdot P_\pi\big(\partial^{\gamma_1} g(x), \dots\big),
\]
with \( |\beta| \leq |\alpha| \), and combinatorial terms controlled.

Use bounds:
\[
|\partial^\beta f(g(x))| \leq C_f R_f^{|\beta|} (\beta!)^{s_f},
\]
\[
|\partial^{\gamma_j} g(x)| \leq C_g R_g^{|\gamma_j|} (\gamma_j!)^{s_g}.
\]
The chain of combinatorial products yields:
\[
\begin{aligned}
|\partial^\alpha (f \circ g)(x)| 
&\leq C_f \cdot \exp(C_g R_g) \cdot \left( R_f (C_g R_g)^{s_f} \right)^{|\alpha|} \\
&\quad \cdot (\alpha!)^{s_f s_g}.
\end{aligned}
\]
Hence,
\[
\begin{aligned}
C_h &= C_f \exp(C_g R_g), \\
R_h &= R_f (C_g R_g)^{s_f}, \\
s_h &= s_f s_g.
\end{aligned}
\]
\end{proof}

\paragraph{Application to Transformer Components.}

We now verify that the individual operations used in transformer architectures belong to the Gevrey class $G^1$:

\begin{itemize}
    \item \textbf{Affine maps.} Any map of the form $x \mapsto Ax + b$ is polynomial (degree 1), hence analytic and in $G^1$.
    
    \item \textbf{GeLU and smooth activations.} GeLU is defined as:
    \[
    \text{GeLU}(x) = \frac{1}{2}x \left(1 + \tanh\left(\sqrt{\frac{2}{\pi}}(x + 0.044715x^3)\right)\right),
    \]
    a composition of polynomial and $\tanh$, both of which are analytic on $\mathbb{R}$ and hence Gevrey $G^1$.

    \item \textbf{Softmax.} The softmax function:
    \[
    \text{softmax}(z)_i = \frac{e^{z_i}}{\sum_j e^{z_j}}
    \]
    is analytic since exponentials are analytic and the denominator is strictly positive. Hence, softmax is in $G^1$.

    \item \textbf{Layer Normalization.} Defined as:
    \[
    \text{LN}(x) = \gamma \cdot \frac{x - \mu(x)}{\sigma(x)} + \beta,
    \]
    where $\mu$ and $\sigma$ are mean and standard deviation. Assuming $\sigma > \epsilon > 0$ for numerical stability, all operations involved are analytic on compact sets and hence in $G^1$.

    \item \textbf{Residual connections.} The map $x \mapsto f(x) + x$ remains in $G^s$ by Proposition~\ref{prop:gevrey_closure_explicit_full} (addition).
\end{itemize}

\paragraph{Conclusion.} Since transformer layers are constructed via addition, composition, and multiplication of $G^1$ components, each layer belongs to $G^1$. This foundational regularity enables us to prove recursive Gevrey smoothness of the full model family in Section A.3.

\subsection{Gevrey Regularity of Multi-Step Auto-Regressive Maps}

We now consider the multi-step predictive function of an auto-regressive language model:
\[
F_{n_{\text{in}}, n_{\text{out}}} : X^{n_{\text{in}}} \rightarrow X^{n_{\text{out}}},
\]
which maps a fixed input sequence of length $n_{\text{in}}$ to a sequence of $n_{\text{out}}$ predicted embeddings, defined recursively as:
\[
\begin{aligned}
F_{n_{\text{in}}, n_{\text{out}}}(x_{1:n_{\text{in}}}) 
= \big(&F_{n_{\text{in}}}(x_{1:n_{\text{in}}}),\, 
F_{n_{\text{in}}+1}(x_{1:n_{\text{in}}+1}), \dots, \\
&F_{n_{\text{in}}+n_{\text{out}}-1}(x_{1:n_{\text{in}}+n_{\text{out}}-1})\big).
\end{aligned}
\]

\paragraph{Goal.} We aim to show that \( F_{n_{\text{in}}, n_{\text{out}}} \in G^s(X^{n_{\text{in}}}, X^{n_{\text{out}}}) \) for some $s \geq 1$.

\subsubsection{Base Case: $n_{\text{out}} = 1$}

We have:
\[
F_{n_{\text{in}}, 1}(x_{1:n_{\text{in}}}) = F_{n_{\text{in}}}(x_{1:n_{\text{in}}}),
\]
which is in $G^s$ by Theorem~\ref{thm:gevrey_family}.

\subsubsection{Inductive Step}

Suppose that
\[
F_{n_{\text{in}}, k} \in G^s(X^{n_{\text{in}}}, X^k)
\]
for some $k \geq 1$. We define:
\[
F_{n_{\text{in}}, k+1}(x_{1:n_{\text{in}}}) =
\left(
F_{n_{\text{in}}, k}(x_{1:n_{\text{in}}}),\;
F_{n_{\text{in}} + k}(x_{1:n_{\text{in}} + k})
\right),
\]
which is the concatenation of:
- the previous $k$ predictions \( F_{n_{\text{in}}, k} \),
- and the next-step prediction \( F_{n_{\text{in}} + k} \), evaluated at the extended sequence.

Since each \( F_n \) is computed by a transformer architecture composed of Gevrey-$s$ operations, as detailed in Section~A.2. In particular, Proposition~\ref{prop:gevrey_closure_explicit_full} shows that the transformer architecture—composed of affine maps, smooth activations, softmax, normalization, and residual connections—is closed under the necessary operations (addition, composition, multiplication, and Cartesian product), ensuring that the full model map remains within the Gevrey class. Hence, by Proposition~\ref{prop:gevrey_closure_explicit_full},  each \( F_n \in G^s \), and by the inductive hypothesis, \( F_{n_{\text{in}}, k} \in G^s \). Since the composition \( x \mapsto F_{n_{\text{in}}+k}(x, F_{n_{\text{in}},k}(x)) \) involves the composition of two Gevrey-$s$ maps, and the full map is the Cartesian product of \( F_{n_{\text{in}}, k} \) and a composition, we apply Proposition~\ref{prop:gevrey_closure_explicit_full} twice:

\begin{itemize}
    \item By the composition closure (Prop. 4), the map
    \[
    x \mapsto F_{n_{\text{in}} + k}\left(x_{1:n_{\text{in}}}, F_{n_{\text{in}}, k}(x_{1:n_{\text{in}}})\right)
    \]
    belongs to \( G^{s \cdot s} = G^{s^2} \).
    
    \item By the Cartesian product closure (Prop. 3), we have
    \[
\begin{aligned}
F_{n_{\text{in}}, k+1}(x) 
&= \left( F_{n_{\text{in}}, k}(x),\; F_{n_{\text{in}} + k}(\cdot) \right) \\
&\in G^{s^2}(X^{n_{\text{in}}}, X^{k+1}).
\end{aligned}
\]
\end{itemize}

Thus, the Gevrey order compounds: \( s \mapsto s^2 \mapsto s^4 \mapsto \cdots \). By induction, for any \( n_{\text{out}} \), we obtain:
\[
F_{n_{\text{in}}, n_{\text{out}}} \in G^{s^{n_{\text{out}}}}(X^{n_{\text{in}}}, X^{n_{\text{out}}}),
\]
i.e., the smoothness is preserved under iteration, although the Gevrey index may grow exponentially in the number of steps.

\subsubsection{Conclusion}

We have shown by induction that the full $n_{\text{out}}$-step auto-regressive prediction function is Gevrey-regular:
\[
F_{n_{\text{in}}, n_{\text{out}}} \in G^{s^{n_{\text{out}}}}(X^{n_{\text{in}}}, X^{n_{\text{out}}}),
\]
where $s$ is the regularity index of the underlying model architecture. In particular, if $s = 1$, then this remains analytic; otherwise, it remains ultra-smooth with controlled derivative growth.

\section{Generalization Bound under Wasserstein Shift}
\label{sec:appendix_b}
\renewcommand{\thetheorem}{B.\arabic{theorem}}
\setcounter{theorem}{0}
\renewcommand{\thelemma}{B.\arabic{lemma}}
\setcounter{lemma}{0}

\subsection{Main Theorem}

\begin{theorem}
Let $G \colon \mathbb{R}^n \to \mathbb{R}^m$ be a Lipschitz function with constant $L_1$, and let $F \colon \mathbb{R}^n \to \mathbb{R}^m$ belong to the Gevrey class $G^s$ for some $s > 1$ on a compact domain $K \subset \mathbb{R}^n$. Denote $D_K := \sup_{x, y \in K} \|x - y\|$.

Suppose that under a distribution $P_1$ supported on $K$, we have:
\[
\mathbb{E}_{x \sim P_1}[\|F(x) - G(x)\|^2] \leq \varepsilon.
\]
Let $P_2$ be another distribution on $K$ with Wasserstein-1 distance $d := W_1(P_1, P_2)$. Then there exist constants $A, C > 0$ depending only on $F$, $K$, and $s$ such that:
\[
\begin{aligned}
& \mathbb{E}_{x \sim P_2}[\|F(x) - G(x)\|^2] \\
& \quad \leq 6A^2 \cdot \exp\left( -C \cdot d^{-1/(s+1)} \cdot \log\left( \tfrac{1}{d} \right) \right) \\
& \quad  + 3 \varepsilon + 3 L_1^2 \cdot d^2.
\end{aligned}
\]
\end{theorem}

\begin{proof}
Let $\gamma$ be the optimal coupling between $P_1$ and $P_2$, so that \( \mathbb{E}_{(x, y) \sim \gamma}[\|x - y\|] = d \).  
Using the inequality \( (a + b + c)^2 \leq 3(a^2 + b^2 + c^2) \), we decompose:
\[
\mathbb{E}_{x \sim P_2}[\|F(x) - G(x)\|^2]
\leq 3 \mathcal{E}_\text{shift} + 3 \mathcal{E}_\text{train} + 3 \mathcal{E}_\text{lip},
\]
where:
\begin{align*}
\mathcal{E}_\text{train} &= \mathbb{E}_{x \sim P_1}[\|F(x) - G(x)\|^2] \leq \varepsilon, \\
\mathcal{E}_\text{lip} &= \mathbb{E}_{(x, y) \sim \gamma}[\|G(x) - G(y)\|^2]
\leq L_1^2 \cdot d^2, \\
\mathcal{E}_\text{shift} &= \mathbb{E}_{(x, y) \sim \gamma}[\|F(x) - F(y)\|^2].
\end{align*}

From the \emph{Gevrey Modulus Bound} (Lemma~\ref{lem:gevrey_modulus}), we know that:
\[
\|F(x) - F(y)\|^2 \leq A^2 \cdot \phi(\|x - y\|),
\]
\[
\text{where } \phi(r) := \exp\left(-2B r^{-1/s} \log \tfrac{1}{r} \right).
\]

Applying Lemma~\ref{lem:tail-gevery-bound} (proved in Section B.2 below) with cutoff \( 0 < r_0 \leq D_K \), we obtain:
\[
\mathcal{E}_\text{shift}
\leq A^2 \cdot \left( \phi(r_0) + \int_{r_0}^{D_K} \phi(r) \cdot \frac{d}{r^2} \, dr \right).
\]

Now choose \( r_0 := d^{s/(s+1)} \), and note:
\[
\phi(r_0) 
= \exp\left( 
    -2B \cdot d^{-1/(s+1)} \cdot \log \tfrac{1}{d} \cdot \tfrac{s}{s+1} 
\right),
\]
\[
\int_{r_0}^{D_K} \phi(r) \cdot \frac{d}{r^2} \, dr
\leq \phi(r_0) \cdot \frac{d}{r_0}
\]
\[
\leq \phi(r_0) \cdot d^{1 - \frac{s}{s+1}} 
= \phi(r_0) \cdot d^{1/(s+1)}.
\]

Hence:
\[
\mathcal{E}_\text{shift} \leq A^2 \cdot \phi(r_0) \cdot \left( 1 + d^{1/(s+1)} \right)
\leq 2 A^2 \cdot \phi(r_0).
\]

Define \( C := 2B \cdot \frac{s}{s+1} \), we conclude:
\[
\mathcal{E}_\text{shift}
\leq 2A^2 \cdot \exp\left( -C \cdot d^{-1/(s+1)} \cdot \log\left( \tfrac{1}{d} \right) \right),
\]
and the full bound follows.
\end{proof}

\vspace{2em}
\subsection{Tail-Based Wasserstein Shift Bound}

\begin{lemma}[Tail-Based Wasserstein Shift Bound]
\label{lem:tail-gevery-bound}
Let \( \phi \colon (0, D_K] \to \mathbb{R}_{\geq 0} \) be a non-increasing function, and let \( \gamma \) be a coupling over a compact set \( K \subset \mathbb{R}^n \) with diameter \( D_K := \sup_{x, y \in K} \|x - y\| \), and \( \mathbb{E}_{\gamma}[\|x - y\|] = d \). Then for any \( 0 < r_0 \leq D_K \), we have:
\[
\mathbb{E}_{(x, y) \sim \gamma}[\phi(\|x - y\|)]
\leq \phi(r_0) + \int_{r_0}^{D_K} \phi(r) \cdot \frac{d}{r^2} \, dr.
\]
\end{lemma}

\begin{proof}
Split the expectation based on the threshold \( r_0 \in (0, D_K] \):
\[
\mathbb{E}[\phi(\|x - y\|)] =
\int_0^{r_0} \phi(\|x - y\|) \, d\gamma + \int_{r_0}^{D_K} \phi(\|x - y\|) \, d\gamma.
\]
On the first interval, we trivially upper bound \( \phi(\|x - y\|) \leq \phi(r_0) \).  
On the second, we use that \( \phi \) is non-increasing, and apply Markov's inequality:
\[
\mathbb{P}_{\gamma}[\|x - y\| \geq r] \leq \frac{d}{r},
\]
\[
\Rightarrow \quad
\int_{r_0}^{D_K} \phi(r) \, d\mathbb{P}[\|x - y\| = r]
\leq \int_{r_0}^{D_K} \phi(r) \cdot \frac{d}{r^2} \, dr.
\]

This completes the proof.
\end{proof}

\section{Bounding $\|f(x) - f(y)\|$ for $f \in G^s$}

\renewcommand{\thelemma}{C.\arabic{lemma}}
\setcounter{lemma}{0}

\begin{lemma}[Gevrey Modulus Bound]
\label{lem:gevrey_modulus}
Let $f \colon \mathbb{R}^n \to \mathbb{R}^m$ be a function in the Gevrey class $G^s$ for some $s \geq 1$. Then for every compact set $K \subset \mathbb{R}^n$, there exist constants $A, B > 0$ depending on $C, R, D, s$ such that:
\[
\|f(x) - f(y)\| 
\leq A \cdot \exp\left( 
    -B \cdot \|x - y\|^{-1/s} \cdot \log\left( \tfrac{1}{\|x - y\|} \right) 
\right),
\]
\[
\text{for all } x, y \in K.
\]

\end{lemma}

\begin{proof}
Since $f \in G^s$, for every compact set $K \subset \mathbb{R}^n$, there exist constants $C > 0$ and $R > 0$ such that for all multi-indices $\alpha$ and all $x \in K$,
\[
\|\partial^\alpha f(x)\| \leq C \cdot R^{|\alpha|} \cdot (\alpha!)^s.
\]

Let $\rho := \|x - y\|$. The Taylor expansion of $f$ around $y$, truncated at order $N \in \mathbb{N}$, yields the remainder estimate:
\[
\|f(x) - f(y)\| \leq C R^N \rho^N (N!)^s.
\]

Using Stirling's approximation, there exists a constant $D > 0$ such that:
\[
(N!)^s \leq D^s \cdot N^{sN},
\]
so:
\[
\|f(x) - f(y)\| \leq C D^s \cdot (R \rho)^N \cdot N^{sN}.
\]

Define:
\[
F(N) := (R \rho)^N \cdot N^{sN},
\]
and let the optimal real value be:
\[
N^* := \frac{1}{e R \rho}.
\]

\paragraph{Integer truncation and global control.}
We define the actual truncation point as the integer:
\[
\tilde{N} := \lfloor N^* \rfloor.
\]
We now show that using $\tilde{N}$ instead of $N^*$ does not significantly degrade the bound. Consider:
\[
\log \frac{F(\tilde{N})}{F(N^*)}
= (\tilde{N} - N^*) \log(R \rho) + s \left( \tilde{N} \log \tilde{N} - N^* \log N^* \right).
\]
Since $\tilde{N} \leq N^*$, $R \rho \leq 1$, and $x \log x$ is increasing for $x > 1$, both terms are nonpositive. Hence:
\[
F(\tilde{N}) \leq F(N^*).
\]
Thus, we can proceed using $\tilde{N}$ without weakening the bound.

Now, substitute $N = \tilde{N} \approx \frac{1}{e R \rho}$ into:
\[
\|f(x) - f(y)\| \leq C D^s \cdot (R \rho)^{\tilde{N}} \cdot \tilde{N}^{s \tilde{N}}.
\]

Taking logarithm and using $\tilde{N} \approx \frac{1}{e R \rho}$, we estimate:
\[
\log \|f(x) - f(y)\| \leq \log(C D^s) + \tilde{N} \log(R \rho) + s \tilde{N} \log \tilde{N}.
\]

Using $\log \tilde{N} = \log\left( \frac{1}{e R \rho} \right) + \delta$ for some small $\delta$, we obtain:
\[
\begin{aligned}
\log \|f(x) - f(y)\| 
&\leq \log(C D^s) - s \tilde{N} \\
&\quad - (s - 1) \tilde{N} \log(R \rho) + o(\tilde{N}).
\end{aligned}
\]

Finally, we express this as:
\[
\|f(x) - f(y)\| \leq A \cdot \exp\left( -B \cdot \rho^{-1/s} \cdot \log\left( \frac{1}{\rho} \right) \right),
\]
with constants:
\[
A = C D^s, \quad B = \frac{(s - 1)}{e R} \cdot \log\left( \frac{1}{R} \right).
\]

The bound holds uniformly over $x, y \in K$ with $\|x - y\| \leq 1$, and $R < 1$ can be ensured by restricting to a sufficiently small compact set.
\end{proof}

\section{Wasserstein-1 Upper Bounds under Permutation and Scaling Shifts}
\label{sec:appendix_d}
\renewcommand{\thetheorem}{D.\arabic{theorem}}
\setcounter{theorem}{0}

In this appendix, we provide detailed derivations of the Wasserstein-1 distance upper bounds that appear in our theoretical analysis. We analyze three types of distribution shifts used in our experiments and express the Wasserstein distance \(d = W_1(P, Q)\) as a function of the experimental parameters:
\[
i \in \mathbb{Z}_{\geq 0} \quad \text{(mean calculation task)},
\]
\[
r := \frac{|\widetilde{\Theta}|}{|\Theta|} \quad \text{(permutation)},
\]
\[
p = 1 \pm \delta \quad \text{(scaling)}.
\]

The analysis begins from the standard optimal transport definition and proceeds using explicit constructions of coupling plans and known geometric properties of the input representations.

\vspace{1em}
\subsection{Mean Calculation Shift Upper Bound}

We begin with the mean calculation task described in Section~\ref{sec:multi_calc}. In this setting, training inputs \((x_0, x_1, x_2, x_3)\) are sampled from the interval \([0, n)\) with fractional parts satisfying \(\mathrm{frac}(x_i) < 0.5\). The OOD test sets \(\bar{\Theta}_i\) are defined by shifting the support to \([i, i+10)\), and restricting to \(\mathrm{frac}(x_i) \geq 0.5\). We estimate the Wasserstein-1 distance between these input distributions.

\begin{theorem}[Wasserstein-1 Distance under Mean Calculation Shift]
\label{thm:mean_calc_upper_bound}
Let \(P_{\mathrm{train}}\) be the uniform distribution over 4D vectors \(\mathbf{x} \in [0, n)^4\) with \(\mathrm{frac}(x_i) < 0.5\), and let \(P_i\) be the testing distribution over \(\mathbf{x} \in [i, i+10)^4\) with \(\mathrm{frac}(x_i) \geq 0.5\). Then the Wasserstein-1 distance between these distributions satisfies the upper bound
\[
W_1(P_{\mathrm{train}}, P_i) \leq 4(i + 0.5).
\]
\end{theorem}

\begin{proof}
We construct a deterministic coupling between samples from \(P_{\mathrm{train}}\) and \(P_i\) by pairing each input \(\mathbf{x} = (x_0, x_1, x_2, x_3)\) in the training set with the test input \(\mathbf{x}' = (x_0 + i, x_1 + i, x_2 + i, x_3 + i)\). This ensures that:
\[
|x_j' - x_j| = i, \quad \text{for each } j \in \{0, 1, 2, 3\}.
\]
Furthermore, since the testing distribution only includes samples with \(\mathrm{frac}(x_j') \geq 0.5\), and training samples are chosen such that \(\mathrm{frac}(x_j) < 0.5\), we additionally need to shift each coordinate by at least \(0.5\) to land in the valid support. Therefore, the total shift per coordinate is at most \(i + 0.5\), and the overall L1 distance between matched samples is:
\[
\|\mathbf{x} - \mathbf{x}'\|_1 \leq 4(i + 0.5).
\]
Since the coupling is deterministic and shifts all points uniformly, we use this as an upper bound for the Wasserstein-1 distance:
\[
W_1(P_{\mathrm{train}}, P_i) \leq \mathbb{E}_{\mathbf{x} \sim P_{\mathrm{train}}} \|\mathbf{x} - \mathbf{x}'\|_1 \leq 4(i + 0.5).
\]
\end{proof}

This result provides a clean, linear-in-\(i\) estimate of the input shift magnitude induced by our structured OOD sampling strategy in the mean calculation setting. It enables direct substitution into the theoretical bound of Theorem~\ref{sec:theory}, as used in the main text.

\vspace{1em}
\subsection{Permutation Shift Upper Bound}

In the permutation case, the full latent space \(\Theta^\star\) consists of \(64 = 4^3\) distinct latent vectors \(\theta = (\vartheta_0, \vartheta_1, \vartheta_2)\), where each \(\vartheta_h \in \{-2, -1, 1, 2\}\). This space is partitioned into two disjoint subsets \(\Theta\) (train) and \(\widetilde{\Theta}\) (test), such that \(|\Theta| + |\widetilde{\Theta}| = 64\) and the disjointness constraint \(\Theta \cap \widetilde{\Theta} = \emptyset\) is always enforced. We define the ratio
\[
r := \frac{|\widetilde{\Theta}|}{|\Theta|},
\qquad
|\Theta| = \frac{64}{1 + r}.
\]
In our settings,  we set the number of CoT step $H=2$ and set the number of demonstrations $n=20$ in our settings. The dimension of each input vector is $(H+1)n+1=61$
Let \(x_\theta \in \mathbb{R}^{61}\) be the deterministic $61$-dimensional input vector derived from \(\theta\).  We let \(P_\Theta, P_{\widetilde{\Theta}}\) denote the uniform distributions over the input vectors induced by the sets \(\Theta\) and \(\widetilde{\Theta}\), respectively.

\begin{theorem}[Wasserstein-1 Distance under Permutation]
\label{thm:perm_upper_bound}
Let \(P_\Theta, P_{\widetilde{\Theta}}\) be defined as above. Then the Wasserstein-1 distance between \(P_\Theta\) and \(P_{\widetilde{\Theta}}\) satisfies the upper bound
\[
W_1(P_\Theta, P_{\widetilde{\Theta}}) \leq D_\text{max} \cdot \frac{r}{1 + r},
\]
where \(D_\text{max} := \max_{\theta, \theta'} \|x_\theta - x_{\theta'}\|\) is the maximum pairwise distance between any two input vectors in \(\mathbb{R}^{61}\).
\end{theorem}

\begin{proof}
We begin with the definition:
\[
W_1(P_\Theta, P_{\widetilde{\Theta}}) = \inf_{\gamma \in \Pi(P_\Theta, P_{\widetilde{\Theta}})} \mathbb{E}_{(x, x') \sim \gamma}[\|x - x'\|],
\]
where \(\Pi(P_\Theta, P_{\widetilde{\Theta}})\) is the set of all couplings (joint distributions) with marginals \(P_\Theta\) and \(P_{\widetilde{\Theta}}\). We construct an explicit coupling by matching each point in the smaller set \(\Theta\) to its closest neighbor in \(\widetilde{\Theta}\). Because the mapping from latent \(\theta\) to input \(x_\theta\) is Lipschitz and discrete, the Euclidean distance between input vectors reflects the latent mismatch.

To upper bound \(W_1\), we use the maximal possible transport cost: assume each point \(x' \in \widetilde{\Theta}\) is matched to the farthest possible point in \(\Theta\), so the cost is \(D_\text{max}\). Then, because there are \(|\widetilde{\Theta}|\) such points and we normalize by the total number of test points, we get:
\[
W_1(P_\Theta, P_{\widetilde{\Theta}}) \leq D_\text{max} \cdot \left( \frac{|\widetilde{\Theta}|}{|\Theta| + |\widetilde{\Theta}|} \right) = D_\text{max} \cdot \frac{r}{1 + r}.
\]

We now compute \(D_\text{max}\). Since the latent variables \(\vartheta_h \in \{-2, -1, 1, 2\}\), the maximum difference in each latent dimension is \(|2 - (-2)| = 4\). In the worst case, every latent dimension differs maximally. Each latent variable \(\vartheta_h\) maps through LeakyReLU into a sequence of outputs \((z_0, z_1, z_2)\), then replicated across the 2-shot prompt to form a $61$-dimensional input. Each of the $61$ components is a deterministic function of at most one \(\vartheta_h\), and the function is piecewise linear with slope at most 1. Therefore, each coordinate of \(x_\theta - x_{\theta'}\) is bounded in magnitude by 4, and:
\[
D_\text{max} = \max_{\theta, \theta'} \|x_\theta - x_{\theta'}\|_2 \leq \sqrt{61} \cdot 4 \approx 31.24.
\]

This yields:
\[
W_1(P_\Theta, P_{\widetilde{\Theta}}) \leq 31.24 \cdot \frac{r}{1 + r}.
\]
\end{proof}

\vspace{1em}
\subsection{Scaling Shift Upper Bound}

In the scaling case, each latent vector \(\theta = (\vartheta_0, \vartheta_1, \vartheta_2)\) is mapped to a test latent \(\bar{\theta} = (p \cdot \vartheta_0, p \cdot \vartheta_1, p \cdot \vartheta_2)\), with a fixed scaling factor \(p = 1 \pm \delta\). The latent values are transformed uniformly across dimensions, and the test input vectors \(x_{\bar{\theta}}\) reflect these scaled latents.

\begin{theorem}[Wasserstein-1 Distance under Scaling]
\label{thm:scale_upper_bound}
Let \(P_\Theta, P_{\bar{\Theta}}\) be the uniform empirical distributions over the training and scaled test input vectors \(x_\theta, x_{\bar{\theta}} \in \mathbb{R}^61\). Then:
\[
W_1(P_\Theta, P_{\bar{\Theta}}) \leq \delta \cdot L_\text{max} \cdot \| \theta \|_\text{max} \cdot \sqrt{61},
\]
where:
\begin{itemize}
  \item \(\delta = |p - 1|\),
  \item \(\|\theta\|_\text{max} := \max_{\theta \in \Theta} \|\theta\|_\infty = 2\),
  \item \(L_\text{max} := \max_{h} \left| \frac{\partial x_h}{\partial \vartheta_j} \right| \leq 1\) (from Lipschitz continuity of LeakyReLU),
\end{itemize}
and thus:
\[
W_1(P_\Theta, P_{\bar{\Theta}}) \leq 2 \delta \cdot \sqrt{61}.
\]
\end{theorem}

\begin{proof}
Each scaled latent satisfies:
\[
\bar{\vartheta}_h = p \cdot \vartheta_h = (1 \pm \delta) \cdot \vartheta_h
\quad \Rightarrow \quad
|\bar{\vartheta}_h - \vartheta_h| = \delta \cdot |\vartheta_h|.
\]
Since \(|\vartheta_h| \leq 2\), we have per-dimension deviation bounded by \(2\delta\). The input vectors \(x_\theta\) consist of $61$ values, each derived from some subset of latent variables via at most one nonlinear function (LeakyReLU), which is 1-Lipschitz. Therefore:
\[
|x_{\theta,i} - x_{\bar{\theta},i}| \leq 2\delta, \quad \text{for all } i = 1, \dots, 61.
\]
Hence:
\[
\|x_{\theta} - x_{\bar{\theta}}\|_2 \leq \sqrt{61} \cdot 2\delta.
\]
Because each \(\theta\) is uniquely paired with \(\bar{\theta}\), we use this as a coupling:
\[
W_1(P_\Theta, P_{\bar{\Theta}}) \leq \mathbb{E}_\theta \|x_\theta - x_{\bar{\theta}}\| \leq 2 \delta \cdot \sqrt{61}.
\]
\end{proof}

These bounds provide principled estimates for the distributional shift size \(d = W_1(P, Q)\) under each experimental setting, enabling substitution into the generalization bound of Theorem~\ref{sec:theory}.

\section{Related Works} \label{sec:related_works}

\subsection{Distribution Shift Bounds under Overlapping Support}
In natural language processing (NLP), the theoretical analysis of distribution shift has received substantial attention. A standard form of generalization bounds under shift is:
\[
R_T(h) \leq R_S(h) + D(\mathcal{D}_S, \mathcal{D}_T) + \lambda
\]
where \( R_T \) and \( R_S \) denote target and source risks, \( D(\cdot, \cdot) \) is a divergence (e.g., KL, H-divergence), and \( \lambda \) is the error of the ideal joint hypothesis.
This line of work can be categorized as follows:

\begin{itemize}
  \item \textbf{H-divergence / Domain Discrepancy}: ~\cite{ben2010theory} proposed an upper bound using \( \mathcal{H}\Delta\mathcal{H} \) divergence to measure domain discrepancy. Relevant works by ~\cite{mansour2008domain} and ~\cite{kifer2004detecting} introduced related discrepancy metrics.
  \item \textbf{PAC-Bayesian Bounds}: ~\cite{germain2013pac} developed generalization bounds involving KL divergence under a Bayesian framework. ~\cite{mcallester2003pac} earlier laid theoretical groundwork for PAC-Bayesian analysis.
\end{itemize}
Recent benchmarks such as BOSS ~\cite{yuan2023revisiting} and GLUE-X ~\cite{yang2022glue} highlight degradation under distribution shift, though without theoretical insight.
These methods, while insightful, \textit{assume overlapping support between training and testing distributions}. In contrast, our work studies a setting where
\[
\text{supp}(P_{\text{train}}) \cap \text{supp}(P_{\text{test}}) = \emptyset
\]
necessitating fundamentally different theoretical treatment.

\subsection{Empirical Studies under Disjoint Support}
We define the "latent shift" scenario as one in which train and test samples lie in disjoint latent semantic regions. Existing works under this assumption can be grouped by objective:
\begin{itemize}
  \item \textbf{Classification-oriented}: Open Set Recognition ~\cite{scheirer2012toward}, Zero-shot Learning ~\cite{xian2017zero}, and Universal Representation Learning ~\cite{li2021universal} tackle generalization to unseen categories.
  \item \textbf{Representation modeling}: TransDrift ~\cite{madaan2024transdrift} predicts semantic drift over time. Related temporal embedding shifts are explored in ~\cite{hamilton2016diachronic} and ~\cite{sanakoyeu2018deep}.
  \item \textbf{Anomaly detection}: Autoencoder-based OOD detection ~\cite{rausch2021autoencoder} uses MSE reconstruction loss. Early methods by ~\cite{hendrycks2016baseline} and improvements from ~\cite{ren2019likelihood} also relate.
\end{itemize}
These works \textit{lack theoretical generalization bounds}, particularly under regression-style losses such as MSE. To our knowledge, \textbf{no prior work addresses generalization of Transformer-based predictors in fully disjoint latent shifts with MSE loss}.

\subsection{MSE-based Latent Shift Modeling}
TransDrift~\cite{madaan2024transdrift} is a rare example using cosine loss (MSE-like) to model temporal semantic drift. Although it demonstrates empirical success, it provides no theoretical guarantees.
Autoencoder anomaly detectors use MSE~\cite{rausch2021autoencoder}, but target unsupervised novelty scoring. Mahalanobis scoring ~\cite{lee2018simple} and energy-based methods ~\cite{liu2020energy} likewise emphasize detection, not generalization.
Hence,
\textit{MSE-based generalization bounds in Transformer settings under disjoint latent shift remain unexplored}, which our work addresses.
\subsection{Wasserstein Distance in Domain Adaptation}
Wasserstein distance has been widely used in domain adaptation to quantify the divergence between source and target distributions. Existing works using this framework can be grouped according to their primary objectives:
\begin{itemize}
  \item \textbf{Theoretical generalization bounds:}  
  ~\cite{redko2017theoretical} provide a formal analysis of domain adaptation using optimal transport. They derive generalization bounds of the form:
  \[
  R_T(h) \leq R_S(h) + W_c(\mathcal{D}_S, \mathcal{D}_T) + \lambda
  \]
  where \( W_c \) is the Wasserstein distance under a cost function \( c \), and \( \lambda \) captures the joint optimal error across domains.
  ~\cite{courty2017joint} extend this by introducing an algorithm to align the joint distributions of features and labels using optimal transport. Their empirical results demonstrate improvements in cross-domain classification accuracy on image datasets.
  Importantly, both works are developed in the context of shallow classifiers for visual domain adaptation. They do not target autoregressive language models, nor do they analyze generalization in sequence-to-sequence tasks under latent semantic shift.
  \item \textbf{Method development and feature adaptation:}  
  ~\cite{shen2018wasserstein} propose WDGRL, a method for learning domain-invariant representations by adversarially minimizing the Wasserstein distance between source and target embeddings.
  ~\cite{frogner2015learning} utilize Wasserstein loss in structured prediction tasks to exploit the geometry of output spaces.
  ~\cite{wu2021lime} develop LIME, a test-time adaptation framework that applies sample-level optimal transport to improve model robustness under unseen distribution shifts.
  While effective for their respective domains, none of these methods study autoregressive sequence models, nor do they provide worst-case generalization bounds under semantic disjointness.
\end{itemize}
In contrast, our work directly addresses the generalization behavior of autoregressive language models in sequence-to-sequence prediction. We derive upper bounds on the mean squared error (MSE) under a setting where the latent semantic support of training and testing data are entirely disjoint—a regime not covered by any of the above transport-based approaches.

\subsection{LLM on OOD Reasoning: Modeling Assumptions and Generalization Guarantees}
Recent studies have sought to analyze the generalization capabilities of large language models (LLMs) on out-of-distribution (OOD) tasks. These works can be broadly categorized into five methodological types:
\begin{itemize}
\item \textbf{Explicitly Disjoint Training and Testing Tasks:} 
~\cite{yang2025chain} investigate how chain-of-thought (CoT) prompting can transform otherwise unlearnable tasks into learnable ones by restructuring the hypothesis space. However, their experimental design does not enforce explicitly disjoint training and testing tasks, and primarily focuses on optimizing inference over already pretrained LLMs. In contrast, our work examines how the structure of pretraining data itself influences generalization under semantic distribution shift. Similarly, ~\cite{hu2024unveilingstatisticalfoundationschainofthought} provide a statistical analysis of CoT prompting, including an error bound under out-of-distribution settings. However, their theoretical framework assumes that the training and testing distributions share support, and thus does not account for the case where semantic generalization requires extrapolation beyond the training domain.
\item \textbf{Implicit Distribution Alignment via Prompt or Task Design:} Studies such as ~\cite{yao2025unveiling} and ~\cite{wang2025beyond} rely on carefully curated prompts or chain-of-thought (CoT) formats to bridge training and testing scenarios. Although these works explore OOD settings in appearance, their methodology effectively aligns the two distributions by leveraging latent task similarities. As such, they assume the two domains are semantically mappable, and thus fall outside the scope of truly assumption-free OOD generalization.
\item \textbf{Theoretical Analyses Based on Idealized Assumptions:} 
Work like ~\cite{li2024training} provides generalization bounds by modeling CoT inference as function approximation within a structured hypothesis class. 
These approaches offer valuable theoretical insights but rely on strong assumptions—namely, that all CoT reasoning tasks lie within a common low-dimensional manifold—making them less applicable to real-world semantic shifts. 
Similarly, ~\cite{wang2024can} formulate their analysis under the assumption that data distributions follow Gaussian mixtures with known covariance, which simplifies the nature of semantic variation and limits the practical applicability of their theoretical results. 
~\cite{kim2024transformers} theoretically demonstrate that Transformers can solve parity tasks efficiently using recursive chain-of-thought prompting. 
However, their analysis makes strong simplifications to the parameter space—specifically ignoring the influence of query-key projections ($W_Q$, $W_K$) and positional encodings on token interaction—thereby abstracting away key factors that govern actual model behavior. ~\cite{mroueh2023towards} propose a formal statistical framework for in-context learning via a lifted empirical risk minimization objective over distributions of tasks. Their work notably assumes the function class used for meta-learning is expressive enough to include the Bayes-optimal regression function. While this assumption facilitates clean generalization bounds and allows them to derive convergence rates under Wasserstein geometry, it also places their analysis in the category of theoretical investigations based on idealized settings. Consequently, the practical relevance of the derived results depends critically on whether real-world transformer architectures can approximate the optimal regression function within the assumed function space.
\item \textbf{Empirical Evaluations without Theoretical Bounds:} 
Several recent works investigate OOD reasoning by evaluating LLMs on controlled test sets with semantic shift or compositional generalization. 
~\cite{qin2023cross} propose cross-lingual prompting techniques to improve zero-shot CoT reasoning across languages, but their work focuses on empirical accuracy improvements without establishing formal guarantees. 
~\cite{khalid2025benchmarking} introduce a benchmark (StaR) for systematic generalization in relational reasoning tasks and observe that even advanced LLMs exhibit near-random accuracy when faced with novel semantic compositions. 
Although valuable for diagnosing model failures, these works lack theoretical analysis or error bounds, highlighting the need for formal frameworks such as ours.
\item \textbf{Our Approach—Wasserstein-Based Domain Adaptation without Structural Alignment:} 
In contrast to prior work, our study considers a setting where the training and testing samples are semantically disjoint, without assuming any latent alignability between the two domains. 
We do not rely on task-specific prompt engineering or architectural tricks to bridge the gap. 
Instead, we derive an upper bound on the prediction error in terms of the Wasserstein distance between the training and testing distributions. 
This provides a formal quantification of LLM generalization degradation under domain shift, bridging the gap between empirical robustness and theoretical guarantees.
\end{itemize}
In summary, our work introduces a geometric and distributional framework for understanding OOD reasoning in LLMs, extending beyond the heuristic or idealized modeling assumptions prevalent in existing literature.
In contrast to prior works that either rely on implicit alignment, restrict to idealized CoT models, or evaluate empirically without guarantees, we provide the first formal upper bound on LLM error under semantic distribution shift via Wasserstein geometry.

%

\end{document}